\documentclass[journal,twoside]{IEEEtran}
\usepackage{times}
\usepackage{epsfig}
\usepackage{graphicx}
\usepackage{amssymb}

\usepackage{tikz}
\usepackage[T1]{fontenc}
\usepackage[font=small,labelfont=bf,tableposition=top]{caption}
\usepackage[font=footnotesize]{subcaption}
\usepackage{color}
\usepackage{bm}
\usepackage{epstopdf}
\usepackage{multicol}
\usepackage{multirow}
\usepackage[numbers,sort&compress]{natbib}
\usepackage[section]{placeins}
\usepackage{url}
\usepackage{verbatim}
\usepackage{algorithm}
\usepackage{algpseudocode}
\usepackage{amsmath}
\usepackage{amssymb}
\usepackage{mathrsfs}

\usepackage{longtable}
\usepackage{rotating}
\usepackage{array}
\usepackage{multicol}
\usepackage{multirow}

\usepackage{flushend}

\ifCLASSINFOpdf
\else
\fi

\DeclareMathOperator*{\argmaxA}{arg\,max} % Jan Hlavacek
\begin{document}
\title{Multiple Instance Choquet Integral Classifier Fusion and Regression for Remote Sensing Applications}

%\begin{comment}

\author{Xiaoxiao~Du,~\IEEEmembership{Student Member,~IEEE,}
        and~Alina~Zare,~\IEEEmembership{Senior Member,~IEEE}% <-this % stops a space
\thanks{X. Du is with the Department
of Electrical and Computer Engineering, University of Missouri, Columbia, MO 65211 USA. E-mail: xdy74@mail.missouri.edu.}% <-this % stops a space
\thanks{A. Zare is with the Department
of Electrical and Computer Engineering, University of Florida, Gainesville, FL 32611 USA. Email: azare@ece.ufl.edu.}% <-this % stops a space
\thanks{This material is based upon work supported by the National Science Foundation under Grant IIS-1723891-CAREER: Supervised Learning for Incomplete and Uncertain Data.}
}

%\end{comment}


%{Shell \MakeLowercase{\textit{et al.}}: Bare Demo of IEEEtran.cls for IEEE %Journals}
%\markboth{}{}%,~Vol.~X, No.~X, X~X }{} %
% Paper headers
\markboth{IEEE TRANSACTIONS ON GEOSCIENCE AND REMOTE SENSING. Preprint Version. Accepted
October 2018}
{Du and Zare: MICI CLASSIFIER FUSION AND REGRESSION FOR REMOTE SENSING APPLICATIONS} 
% Use only for final version




\maketitle

\begin{abstract}
In classifier (or regression) fusion the aim is to combine the outputs of several algorithms to boost overall performance.   Standard supervised fusion algorithms often require accurate and precise training labels.  However, accurate labels may be difficult to obtain in many remote sensing applications. This paper proposes novel classification and regression fusion models that can be trained given ambiguously and imprecisely labeled training data in which training labels are associated with sets of data points (i.e., ``bags'') instead of individual data points (i.e., ``instances'') following a multiple instance learning framework.  Experiments were conducted based on the proposed algorithms on both synthetic data and applications such as target detection and crop yield prediction given remote sensing data. The proposed algorithms show effective classification and regression performance. 
\end{abstract}


% \fi
\IEEEpeerreviewmaketitle

\section{Introduction}
\label{sec:intro}
Classifier fusion methods aim to combine and integrate multiple classifier outputs while reducing uncertainties in the data, providing more detailed
information and more accurate prediction \cite{hackett1990multi1, zhang2010multi}.  Each of the classifier (or regressor) outputs to be fused may provide complementary or reinforcing information that is helpful for a specific target detection, classification, or regression application \cite{gader2004multi}.

\begin{figure}[h]
%\raggedleft
\begin{subfigure}[h]{0.32\linewidth}
\includegraphics[scale=0.25, width=\textwidth,trim={7mm 15mm 7mm 12mm},clip]{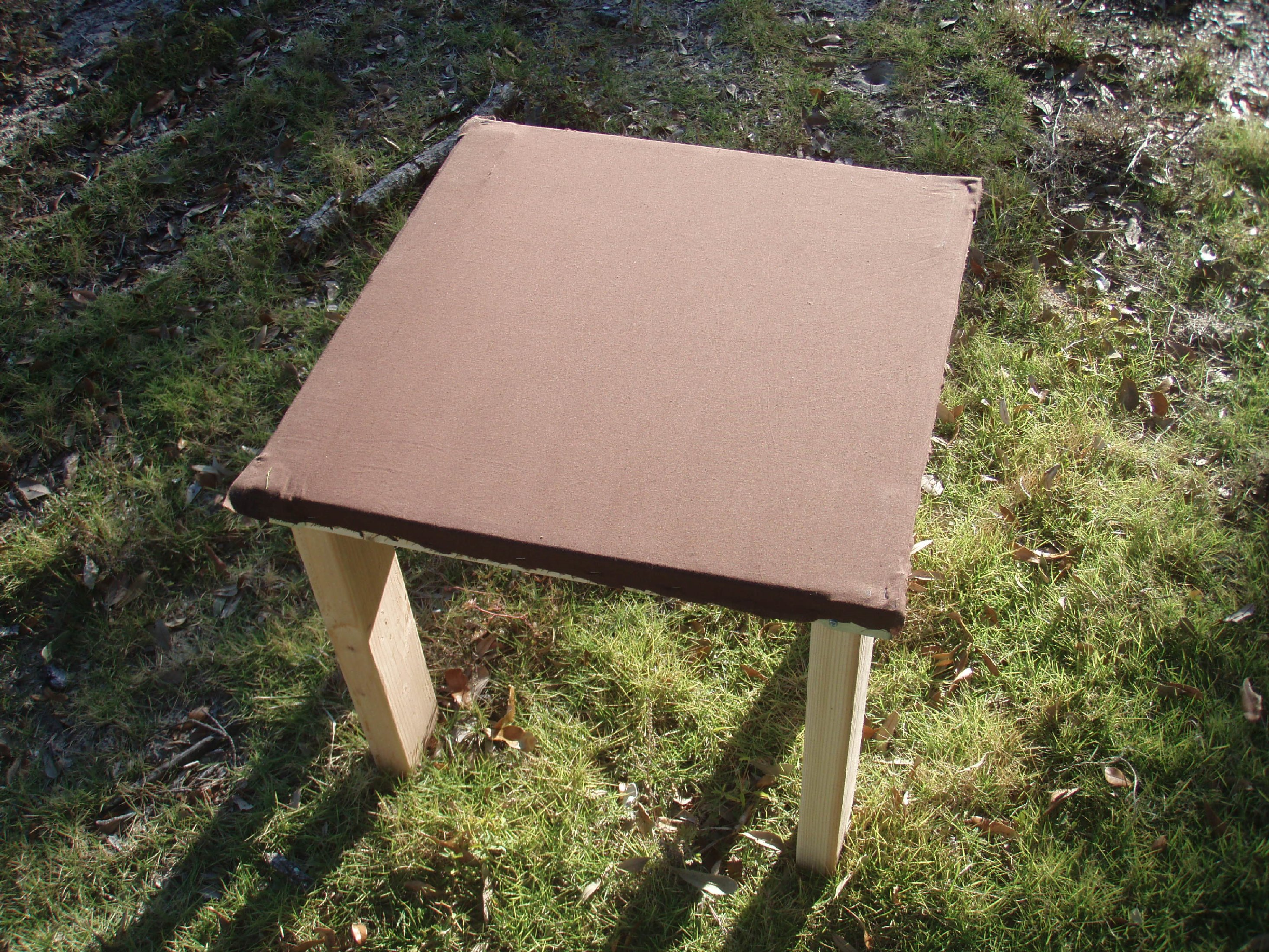}
\caption{ }
\label{fig:browntgt_1m}
\end{subfigure}
\begin{subfigure}[h]{0.32\linewidth}
\includegraphics[scale=0.25, width=\textwidth,trim={7mm 15mm 7mm 12mm},clip]{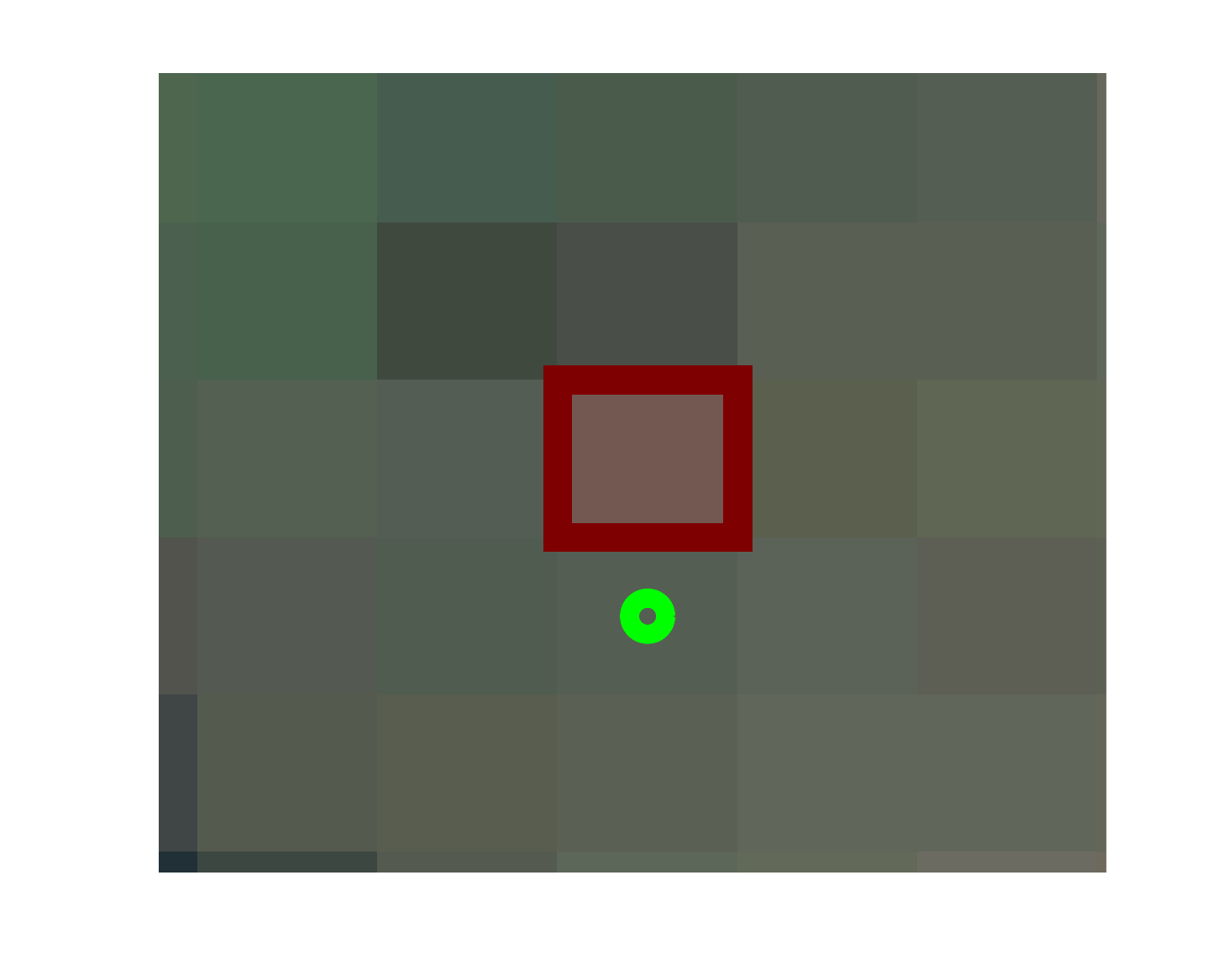}
\caption{ }
\label{fig:browntgt_illu1}
\end{subfigure}
\begin{subfigure}[h]{0.32\linewidth}
\includegraphics[scale=0.25, width=\textwidth,trim={7mm 15mm 7mm 12mm},clip]{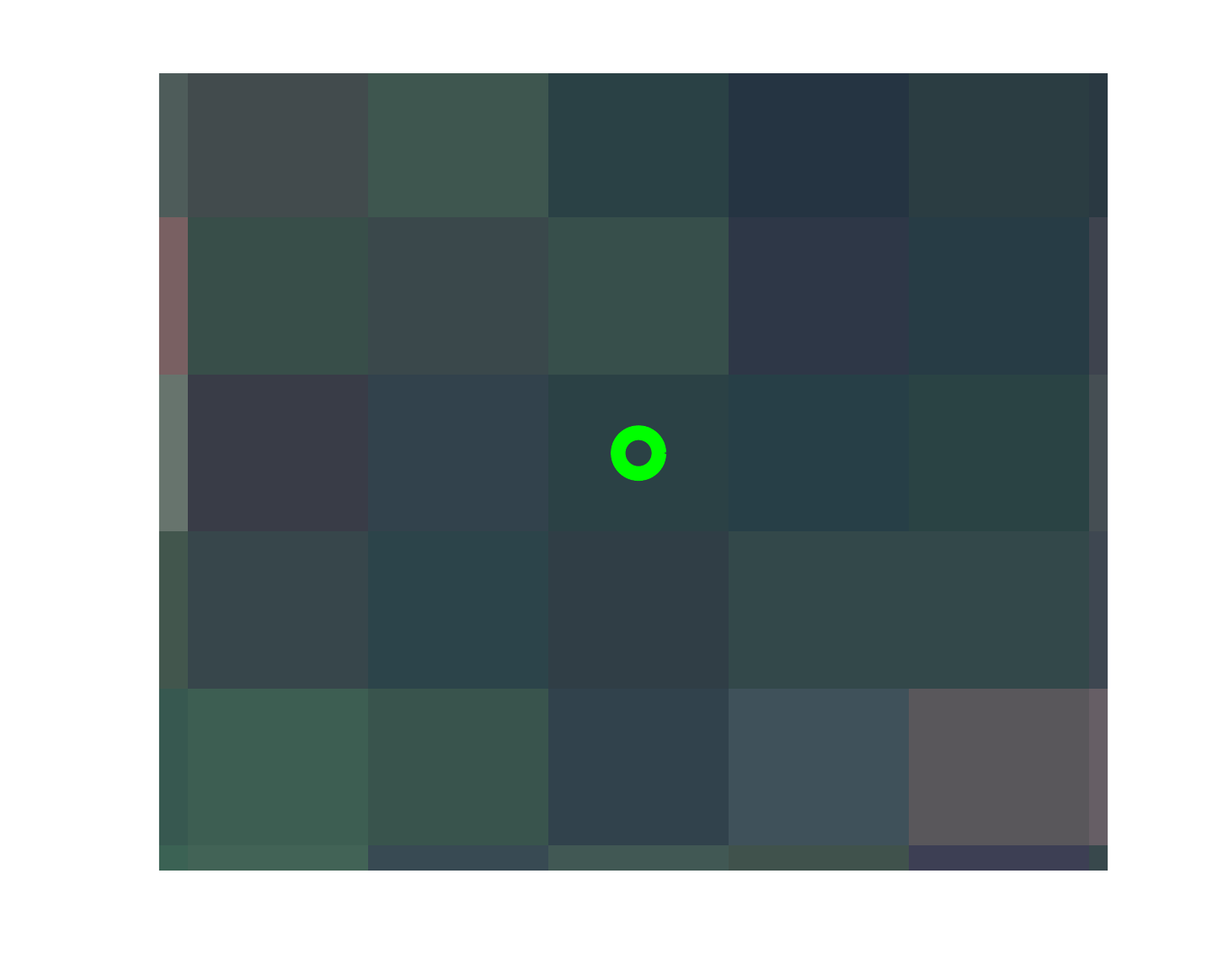}
\caption{ }
\label{fig:browntgt_illu3}
\end{subfigure}
\begin{subfigure}[h]{0.32\linewidth}
\includegraphics[scale=0.25, width=\textwidth,trim={7mm 15mm 7mm 12mm},clip]{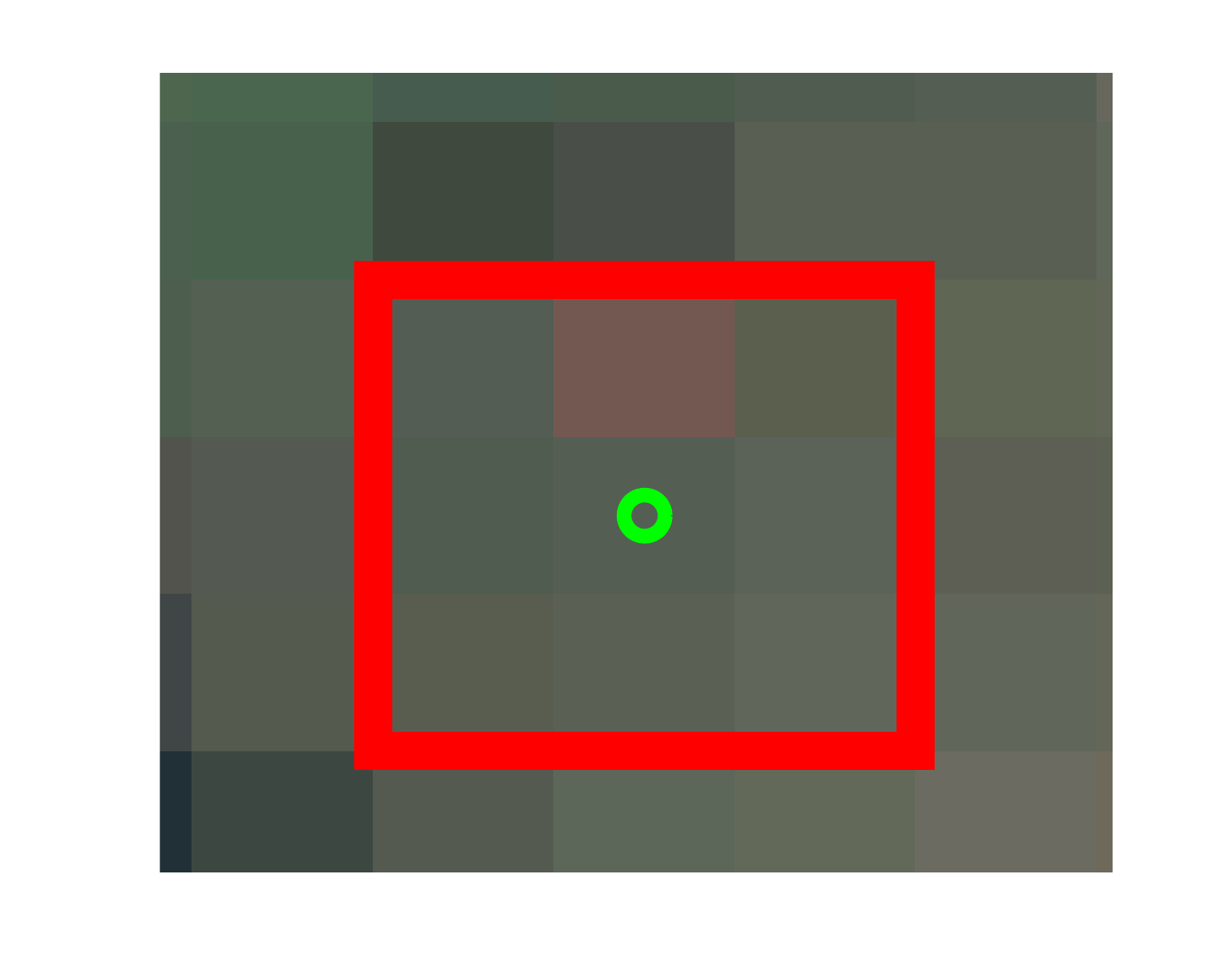}
\caption{ }
\label{fig:browntgt_illu2}
\end{subfigure}
\begin{subfigure}[h]{0.32\linewidth}
\includegraphics[scale=0.25, width=\textwidth,trim={7mm 15mm 7mm 12mm},clip]{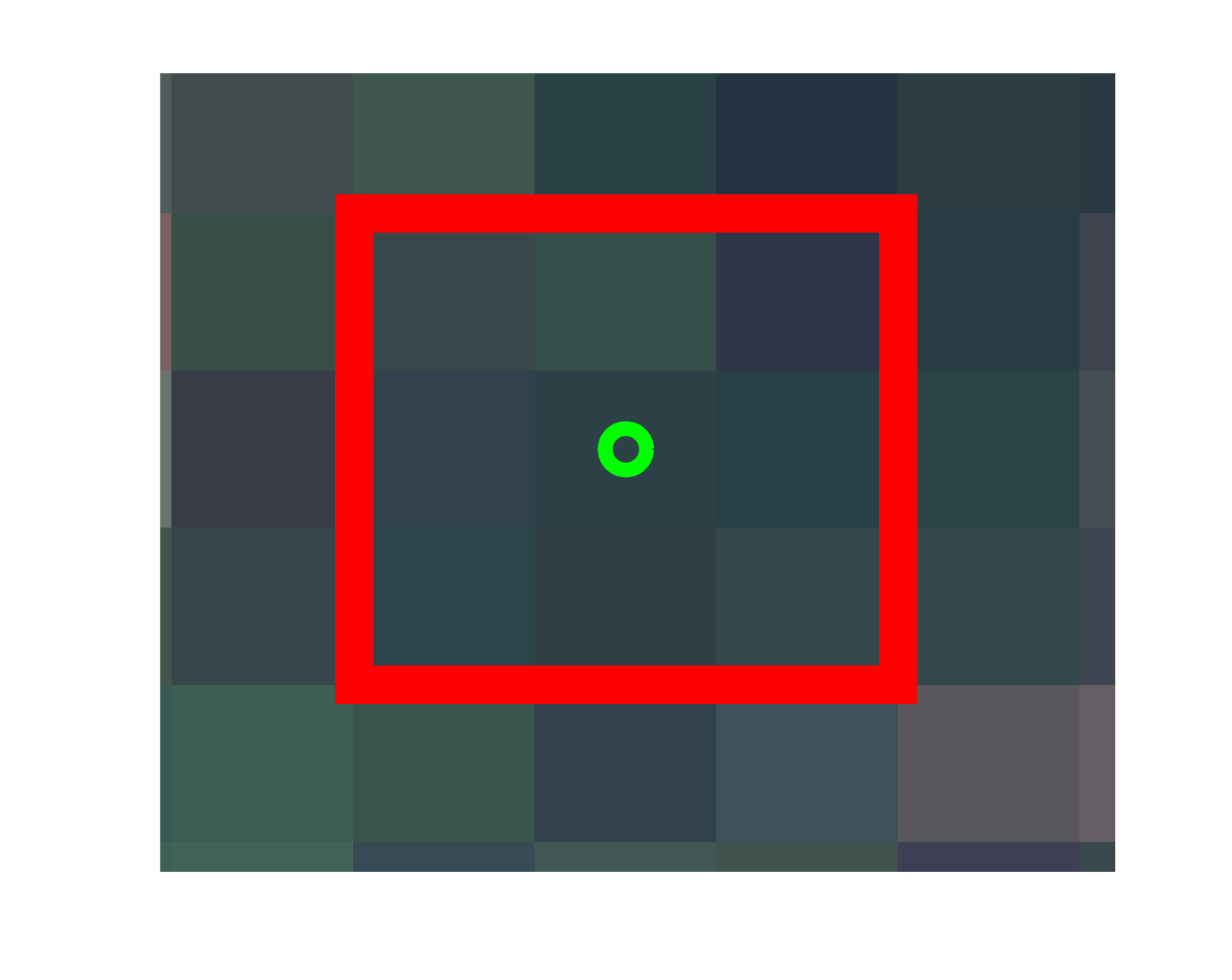}
\caption{ }
\label{fig:browntgt_illu4}
\end{subfigure}
\caption{Illustration of a target application given remote sensing data. (a) A brown target. (b) An inaccurate GPS location (green circle) that differs from the true brown target location (dark red rectangle) in the scene. (b) A sub-pixel target that cannot be visibly seen from the imagery (occluded by vegetation).  (c)(d) Red rectangles mark the approximate region that contains the targets.}
\label{fig:browntgt_illu}
\end{figure}

Previous supervised fusion algorithms developed for remote sensing data often require precise labels for each training data point \cite{waske2007fusion, du2009hyperspectral, yang2010decision, frigui2010context}. However, data-point specific labels are often either unavailable, difficult or expensive to obtain in remote sensing applications. For example, consider the following target detection problem. We have collected hyperspectral (HSI) imagery over the University of Southern Mississippi-Gulfpark campus and the goal is to find targets emplaced within the scene \cite{gader2013muufl}. The ground sample distance of the obtained HSI imagery is $1m$. This is a challenging target detection problem in which many targets are occluded (hidden under a tree, for example) and many are sub-pixel (target size less than $1m^2$). A Global Positioning System (GPS) device was used to obtain target coordinates during data collection. However,  the GPS device used was only accurate to the level of several pixels. In this problem, we know the approximate locations of the targets given the GPS information but we cannot pin-point the exact target locations. Figure~\ref{fig:browntgt_illu} shows an example of remote sensing data burdened with such imprecise and uncertain labels. 

%targets in this scene. Figure~\ref{fig:browntgt_illu1} shows the RGB image over a brown target (the brown colored pixels in the scene). The red cross marks the target coordinates collected by the (inaccurate) GPS device. As can be seen, the GPS location is visibly different than where the brown target actually is in the scene. Figure~\ref{fig:browntgt_illu3} shows a sub-pixel target in the scene hidden under a tree. As can be seen, the pixels nearby are all in green color (color of the leaves of the tree) and the target is not visible in the scene. It would be impossible for humans to visibly locate and manually label where the sub-pixel target is in the scene. For both cases, it is difficult, or impossible, to obtain the accurate target locations in the scene. 

In the above example, it is possible to identify an approximate region (a set of pixels) that contain the targets, as shown in Figure~\ref{fig:browntgt_illu2} and Figure~\ref{fig:browntgt_illu4}. We do not know which pixel(s) exactly correspond(s) to the target, but we know the target is somewhere in the red rectangle region. That is to say, the accurate target location per pixel is not available, but the regions or sets of pixels that contain the targets can be easily obtained from the (imprecise) GPS coordinates. It would be useful, therefore, to develop a trained classifier fusion and regression method that can be trained on labels with such uncertainty in order to better perform target detection and many other tasks given remote sensing data.

The Multiple Instance Learning (MIL) framework was first proposed in \cite{dietterich1997solving} to address uncertainty and inaccuracy in labeled data in supervised learning applications. Since then, the MIL has been used in various remote sensing applications such as spatial-spectral classification \cite{liu2018deep}, landmine detection \cite{yuksel2015multiple, manandhar2015multiple}, seabed segmentation \cite{du2015possibilistic, Cobb2017Multiple}, and target signature characterization \cite{jiao2015functions, zare2017discriminative}. However, the majority of these studies were only applied to single-sensor classification problems. On the other hand, previous supervised fusion methods \cite{waske2007fusion, du2009hyperspectral, yang2010decision, frigui2010context} often require accurate pixel-wise training label and can not handle uncertain/imprecise labels. There has been relatively little work on applying MIL to the label uncertainty problem in decision-level fusion. This paper aims at addressing the problem uncertain/imprecise training labels for multi-sensor/multi-classifier fusion applications by formulating the fusion problem as an MIL problem.

We use the Choquet integral (CI) as the aggregation operator for our proposed fusion framework. The Choquet integral (CI) has a long history of being an effective aggregation operator for non-linear fusion \cite{marichal2000an, labreuche2003thechoquet, mendezvazquez2008minimum}. Compared with commonly used aggregation operators such as weighted arithmetic means \cite{fodor1995characterization}, the Choquet integral is able to model the relationship amongst the combinations of fusion sources and can flexibly represent a wide variety of aggregation operators \cite{maron1998phdthesis, marichal2000an}. The Sugeno integral \cite{Sugeno74} is also a flexible, non-linear aggregator. In this work, we focus on the use of the Choquet integral since, in some cases in the literature, it has shown better classification performance compared with the Sugeno integral \cite{gader2004multi, martinez2016comparison}. However, certainly future work will involve developing MIL Sugeno-based fusion approaches for comparison.  

Standard CI fusion has been previously explored in remote sensing applications \cite{gader2004multi, gader2004continuous, scott2018enhanced, zhao2008object}. However, these methods require accurate per-pixel training labels, which are difficult or impossible to obtain in remote sensing applications, as discussed above. We previously proposed a noisy-or approach that can perform CI fusion with uncertain labels \cite{du2016multiple}, but it only works specifically for two-class classification. It would be useful, therefore, to develop an CI fusion method that can handle uncertain labels for both classifier fusion and regression in remote sensing applications.

This paper proposes a Multiple Instance Choquet Integral (MICI) framework for both multi-sensor classifier fusion and regression that can learn from ambiguously and imprecisely labeled training data. The proposed MICI fusion methods follow the Multiple Instance Learning (MIL) framework and performs fusion using the Choquet integral given uncertain and imprecise training labels. 
Two novel MICI classifier fusion models, the min-max model and the generalized-mean model, are proposed. The classifier fusion models are useful for target detection given remote sensing data, among many other classification tasks. This paper also proposes a Multiple Instance Choquet Integral Regression (MICIR) model for regression problems where the desired prediction is real-valued. The proposed model can fuse multiple sources with real-valued label as well as handling the uncertainties in the training labels. %A monotonic normalized fuzzy measure is learned to be used with the Choquet integral to perform two-classs classifier fusion given bag-level training labels. An optimization scheme using an evolutionary algorithm is used to optimize the models proposed. %Comprehensive experiments were conducted on the proposed models based on the recently proposed MICI noisy-or model \cite{du2016multiple}, and t

 %The proposed model was applied to real regression tasks such as predicting crop yield given remote sensing data.

% The model adopts a ``primary-instance'' assumption that there is one primary instance responsible for the label for each bag \cite{ray2001multiple}. 

The paper is organized as follows. Section~\ref{sec:intro} introduces the problem of uncertain labels in remote sensing data and describes our motivation in using the MIL framework and the CI fusion, which are the basis of the proposed algorithms. Section~\ref{sec:relatedwork} describes previous work in MIL and CI fusion, particularly in remote sensing applications. Section~\ref{sec:algorithm} introduces the proposed classifier fusion and regression models and Section~\ref{sec:optimization} describes the optimization approach.  Section~\ref{sec:experiments} presents fusion results on real target detection and crop yield prediction applications given remote sensing data. Section~\ref{sec:conclusion} provides a summary and conclusion of the work.

\section{Related Work}
\label{sec:relatedwork}
This section introduces related work in multiple instance classification, classifier fusion, and multiple instance regression. This section also describes in detail the previously proposed noisy-or classifier fusion model to be compared with the newly proposed algorithms.
\subsection{Multiple Instance Classification}

\begin{figure}[h]
\hspace{-7mm}
\hspace{5mm}
\begin{minipage}[t]{0.75\linewidth}
\centering
%\hspace{20mm}

\resizebox{2cm}{!}{
\includegraphics[width=0.4\textwidth]{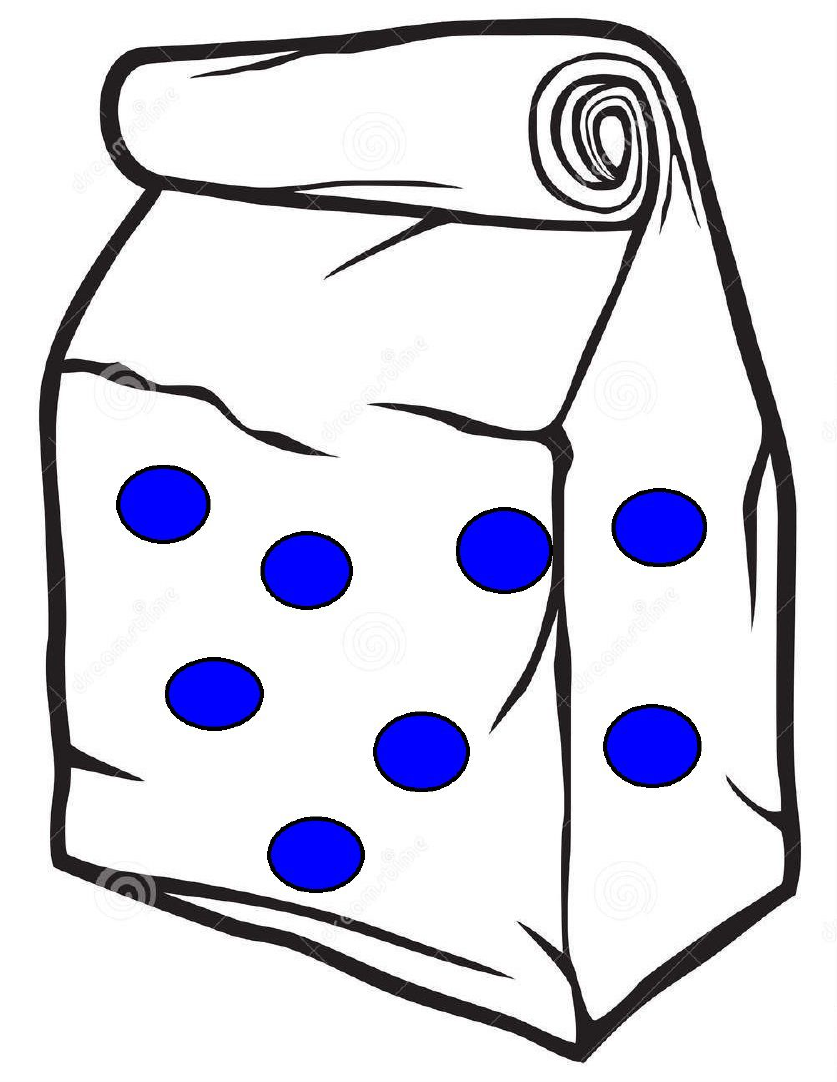}
}
\resizebox{2cm}{!}{
\includegraphics[width=0.4\textwidth]{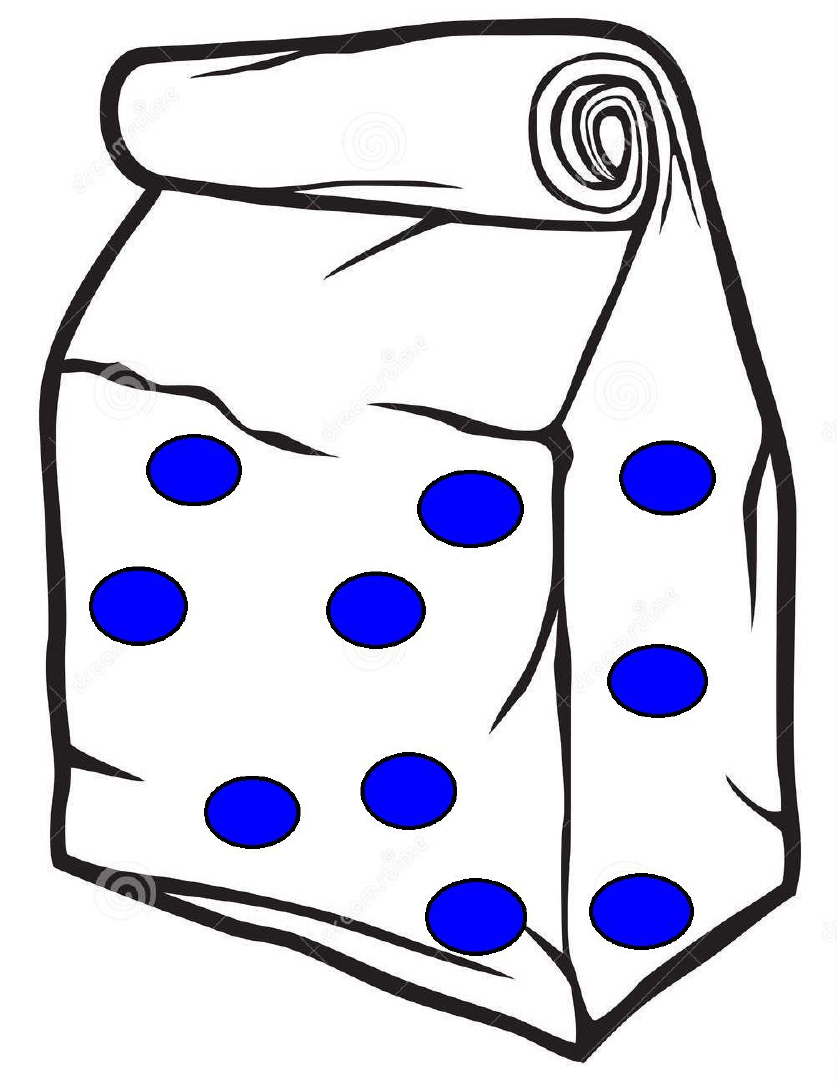}
}

\vspace{5mm}

\resizebox{2cm}{!}{
\includegraphics[width=0.4\textwidth]{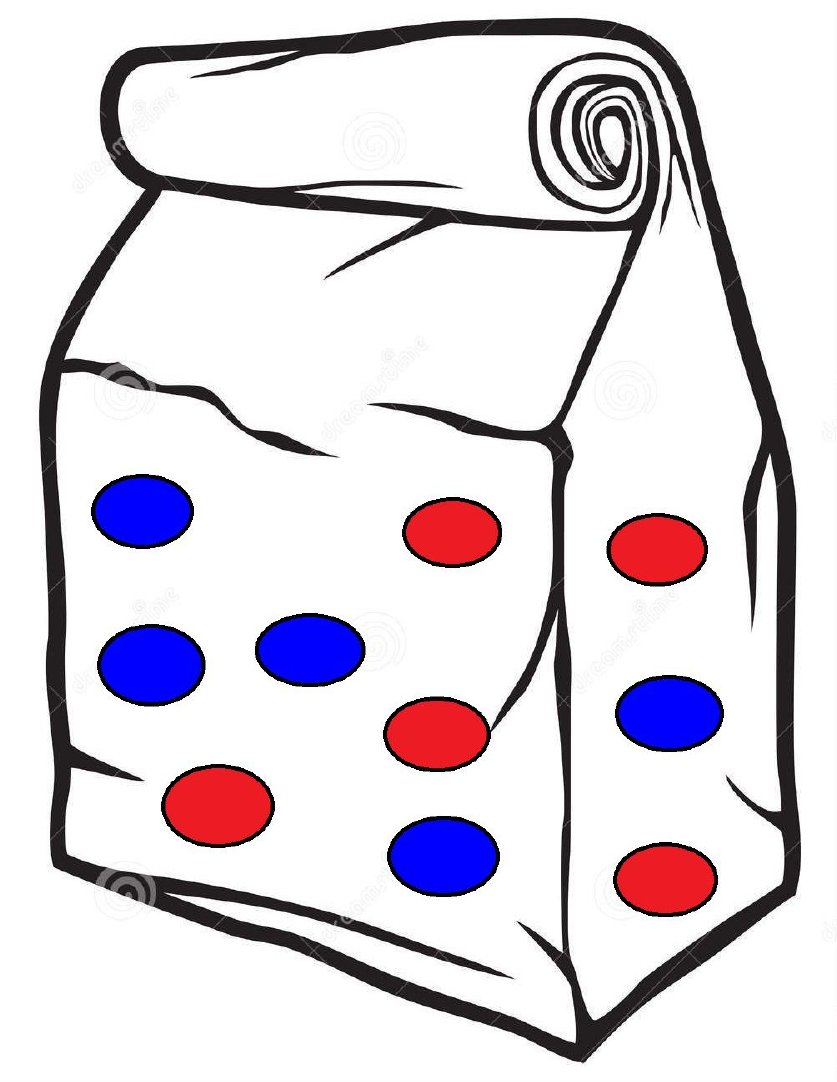}
}
\resizebox{2cm}{!}{
\includegraphics[width=0.4\textwidth]{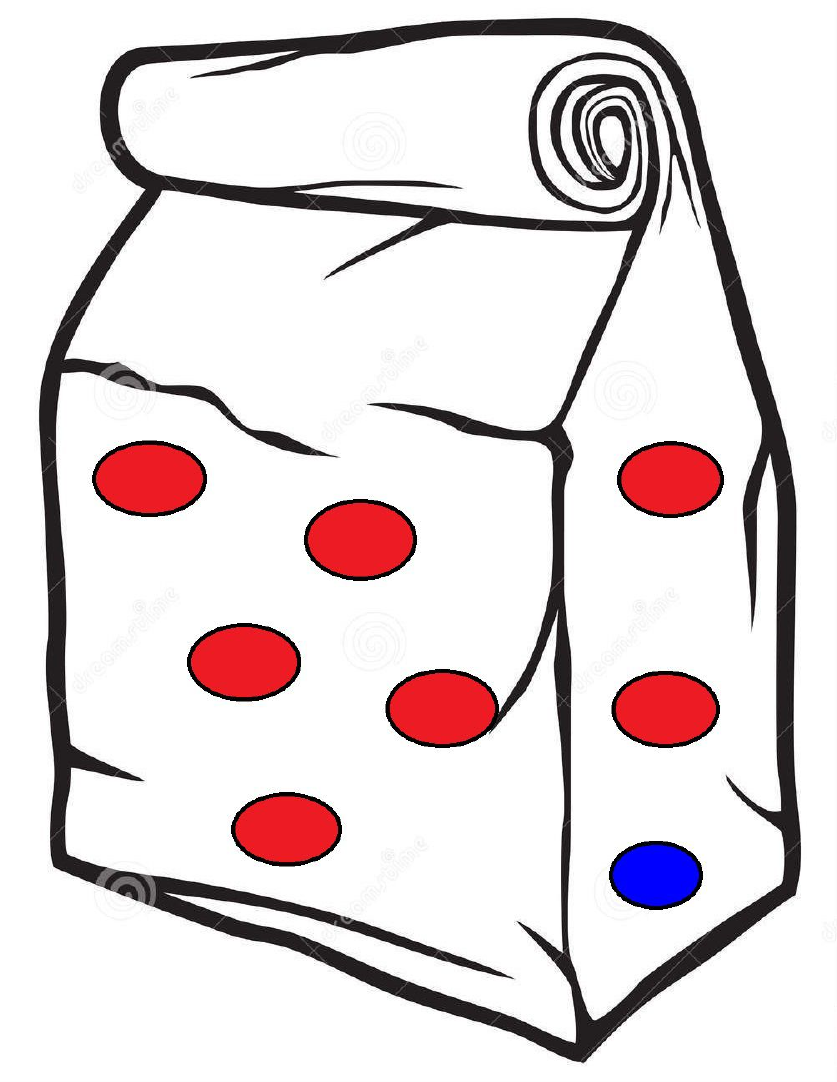}
}
\resizebox{2cm}{!}{
\includegraphics[width=0.4\textwidth]{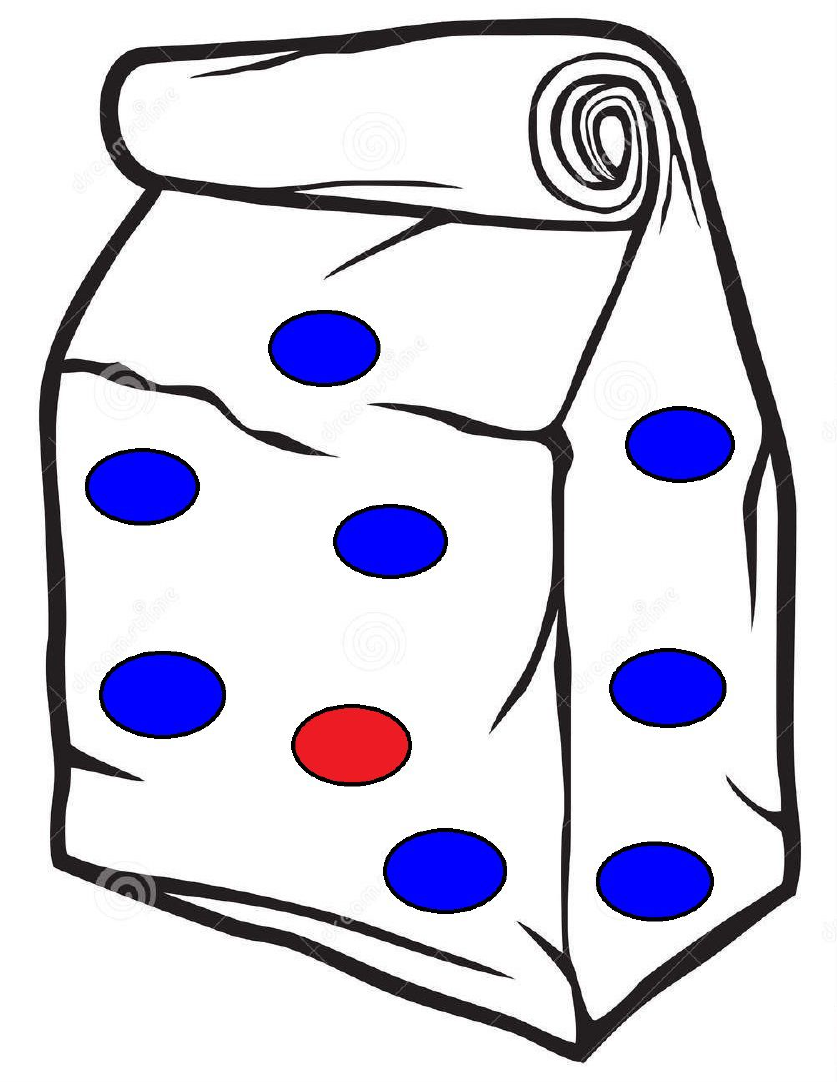}
}
\end{minipage}
\begin{minipage}[t]{0.2\linewidth}
\centering

\vspace{-18mm}

\textbf{\textcolor{blue}{Negative Bags}}

\vspace{18mm}

\textbf{\textcolor{red}{Positive Bags}}
\end{minipage}
  \caption[Illustration of bags in multiple instance learning.]{Illustration of bags in multiple instance learning. Red color marks positive instances and blue color marks negative instances. The two bags on the top row are negative bags and the three bags on the bottom row are positive bags.}
  \label{fig:miltraintestbags}
\end{figure}

In the MIL framework, training labels are associated with sets of data points (``bags'') instead of each data point (``instance''). In the scenario of two-class classification, the standard MIL assumes that a bag is labeled positive if at least one instance in the bag is positive and a bag is labeled negative if all the instances in the bag are negative. Figure~\ref{fig:miltraintestbags} shows an illustration of MIL bags.

\begin{comment}
Figure~\ref{fig:miltraintest} illustrates the differences between standard supervised classification and multiple instance learning classification based on the paradigms discussed in \cite{amores2013multiple}. The standard supervised classification (the instance-space paradigm) learns a classifier based on a set of training feature vectors, where each feature vector has an associated class label (such as marked in orange or blue color in Figure~\ref{fig:miltraintest}) \cite{amores2013multiple}. The instance-space paradigm classifies instances based on their values at the instance level and draws a decision boundary between classes. The MIL classification (the bag-space paradigm), on the other hand, discriminates information at the bag-level. Distances between bags are computed and a standard distance-based classifier may be applied such as the $K$-nearest neighbor (KNN) classifier \cite{peterson2009k, amores2013multiple}. 

\begin{figure}[h]
\centering
  \includegraphics[width=3.5in, trim=0cm 5cm 0cm 0cm, clip=true]{Figures/mil.pdf}
  \caption{Illustration of standard supervised classification and multiple instance learning classification. The left column is training stage and the right column is testing stage. Orange color marks positive instances/bags and blue color marks negative instances/bags in the feature space.}
  \label{fig:miltraintest}
\end{figure}
\end{comment}

The EM-DD and mi-SVM algorithms are two widely cited multiple instance learning techniques for classification \cite{xu2017weakly, quellec2017multiple, zare2017discriminative, cao2017weakly}. The two algorithms are used later in this paper as comparison MIL approaches. The EM-DD algorithm \cite{zhang2001dd}  combines the Expectation-Maximization (EM)  \cite{dempster1977maximum} with the Diverse Density (DD) objective function \cite{dietterich1997solving, maron1998phdthesis, maron1998aframework} to address MIL two-class classification problems.  EM-DD views the relationship between all the instances in the bag and the label of the bag as a latent variable that can be estimated using the EM approach \cite{zhang2001dd}. In the E-step, one instance is picked from each bag as the most influential instance for its bag-level label. In the M-step, the DD is maximized by a gradient ascent search. The process is iterated until a stopping criteria is met.

The mi-SVM algorithm was proposed by Andrews et al.  \cite{andrews2002support} as an MIL extension to support vector machine (SVM) learning approaches. The mi-SVM algorithm can work with bag-level labels in MIL. For two-class classification problems, the mi-SVM objective function is defined as  \cite{andrews2002support}:
\begin{equation}
\min_{\{y_i\}} \min_{\mathbf{w},b,\xi } \frac{1}{2} \left \| \mathbf{w} \right \|^2 + C \sum_i \xi_i,
\end{equation}
such that
\begin{equation}
\sum_{i \in I} \frac{y_i+1}{2} \geq 1, \\
\forall i: y_i(<\mathbf{w}, \mathbf{x}_i> + b) \geq 1- \xi_i, \xi_i \geq 0
\end{equation}
where $\mathbf{w}$ is the weights, $b$ is the bias, $y_i$ is the instance-level label, $\xi_i$ are the slack variables (similar to that of a standard SVM).  The label vector $\mathbf{y}$ represents bag-leveled labels that satisfy the MIL assumption. The mi-SVM algorithm learns a linear discriminate function and maximizes the margin to separate the positive from the negative classes based on instance labels.

\subsection{Multiple Instance Regression}

Multiple instance regression (MIR) addresses multiple instance problems where the prediction values are real-valued.  MIR has been used in the literature for applications such as predicting the ability of antigen peptides to bind to major histocompatibility complex class II (MHC-II) molecules \cite{EL2011predicting}, predicting aerosol optical depth from remote sensing data \cite{wang2008aerosol,wang2012mixture}, and predicting crop yield \cite{wagstaff2007salience,wagstaff2008multiple,wang2012mixture}.

Ray and Page \cite{ray2001multiple} first proposed an MIR algorithm based on the ``primary-instance'' assumption, which assumes there is one primary instance in a bag that is responsible for the real-valued bag-level label. Wagstaff et al. then investigated using MIR in predicting crop yield from real remote sensing data set \cite{wagstaff2007salience, wagstaff2008multiple}. They provided an MI-ClusterRegress algorithm (or in some references, Cluster MIR algorithm) that mapped instances onto (hidden) cluster labels \cite{wagstaff2008multiple}. However, the Cluster MIR algorithm works under the assumption that the instances from a bag are drawn (with noise) from a set of underlying clusters and one of the cluster must be ``relevant'' to the bag-level labels. The algorithm assumes that the  bag data follows a mixture of Gaussian distributions \cite{wagstaff2008multiple}.  More recently, Trabelsi and Frigui \cite{trabelsi2018robustthesis} proposed a Robust Fuzzy Clustering for MIR (RFC-MIR) algorithm that, similar to Cluster-MIR, clusters the training instances and learns multiple regression models for each cluster. Both Cluster-MIR and RFC-MIR pool all instances from all training bags in order to perform clustering.

\subsection{Choquet Integral and Fuzzy Measures}
\label{sec:ciforcf}
%Decision-level fusion takes multiple classifier or real-valued prediction outputs estimated from the data as sources of information. It then derives one fused output with the aim of providing a finer, more accurate and/or descriptive classification or regression result \cite{ruta2000overview, aue2002generalized, du2016multiple}. Classifier fusion has wide applications including remote sensing and hyperspectral image classification \cite{waske2007fusion,du2009hyperspectral,yang2010decision},  object detection \cite{mahmoudi2015object}, landmine detection \cite{keller2000new, keller2001experiments, gunatilaka2001feature, aue2002generalized, frigui2010context}, handwriting recognition \cite{gader1996fusion}, and medical and fault diagnosis \cite{zhang2014research, khazendar2014automated, zhang2010research}.
% A discrete Choquet integral integrates the input sources with respect to a fuzzy measure \cite{keller2016fundamentals}. 

The Choquet integral (CI) is an aggregation operator based on the fuzzy measures \cite{keller2016fundamentals}. Depending on the values of each element in the fuzzy measure, the CI can represent a variety of relationships and combinations amongst the information sources. Therefore, a crucial aspect of using the CI for information/sensor fusion is learning the fuzzy measures for the CI \cite{anderson2010learning, anderson2014regular}. The algorithms proposed in this paper rely on the Choquet integral and fuzzy measures and this subsection provides the definition of the Choquet integral and fuzzy measures.

Consider the case that there are $m$ sources, $C=\left\{c_1,c_2,\dots,c_m\right\}$, for fusion. The ``sources'' refer to the outputs from the set of $m$ classifiers or regressors to be fused. The set $C$ contains $2^m-1$ non-empty subsets. The power set of all (crisp) subsets of $C$ is denoted $2^C$. A monotonic and normalized fuzzy measure, $\mathbf{g}$, is a real valued function that maps $2^C \rightarrow [0, 1]$. It satisfies the following properties \cite{choquet1954theory,Sugeno74, fitting2003beyond, mendezvazquez2008info}:

\indent \indent 1. $g(\emptyset) = 0$;

\indent  \indent  2. $g(C) = 1$; \textit{normalized}

\indent  \indent 3. $g(A) \leq g(B)$ if $A \subseteq B$ and $A, B \subseteq C$.  \textit{monotonic}

In this paper, the monotonic normalized discrete Choquet integral is used as the aggregation operator for both classifier fusion and regression. Suppose we are fusing $m$ classifier or regressor sources. Denote the classifier/regressor output of $k^{th}$ classifier/regressor, $c_k$, on $n^{th}$ data point/instance, $\mathbf{x}_n$, as $h(c_k; \mathbf{x}_n)$. The discrete Choquet integral on instance $\mathbf{x}_n$ given $C$ is then computed as \cite{mendezvazquez2008info, keller2016fundamentals, du2016multiple}:
\begin{equation}
C_\mathbf{g}(\mathbf{x}_n) = \sum_{k=1}^{m}\left[ h(c_k; \mathbf{x}_n) - h(c_{k+1}; \mathbf{x}_n)\right]  g(A_k),
\label{eq:cg}
\end{equation}
where $C$ is sorted so that $h(c_1; \mathbf{x}_n)\geq h(c_2; \mathbf{x}_n) \geq \dots \geq h(c_m; \mathbf{x}_n)$. Since there are only $m$ sources, $h(c_{m+1}; \mathbf{x}_n)$ is defined to be zero. The fuzzy measure element value corresponding to the subset $A_k = \left\{ c_1, \dots, c_k\right\}$ is $g(A_k)$. 

%The Choquet integral has been used effectively in applications such as landmine detection \cite{gader2001recognition, gader2004multi, keller2001experiments, zare2008vegetation}, remote sensing image classification \cite{wang2015integration}, and fall detection \cite{liu2012fall}.

%\subsection{Previously Proposed MICI Noisy-or Model for Multiple Instance Classifier Fusion}
%\label{sec:micinoisyor}

Previously, we proposed a Multiple Instance Choquet Integral classifier fusion method that also uses the Choquet integral and multiple instance learning framework with a noisy-or objective function \cite{du2016multiple}. Recall that in standard MIL, a bag is labeled negative if all the instances in the bag are negative and a bag is labeled positive if there is at least one positive instance in the bag. We modeled the MIL assumption using a noisy-or objective function \cite{maron1998aframework, du2016multiple}:
\begin{equation}
  \begin{aligned}
\ln p(\mathbf{X} |\boldsymbol{\theta}) &= \sum_{a=1}^{B^-} \sum_{i=1}^{N^-_b} \ln \left( 1 - \mathscr{N}\left( C_\mathbf{g}(\mathbf{x}_{ai}^-) | \mu, \sigma^2 \right) \right) \\
&+  \sum_{b=1}^{B^+}   \ln\left( 1 -\prod_{j=1}^{N^+_b}  1- \mathscr{N}\left( C_\mathbf{g}(\mathbf{x}_{bj}^+) | \mu, \sigma^2 \right) \right),
\label{eq:objln}
\end{aligned}
\end{equation}
where $B^+$ is the total number of positive bags, $B^-$ is the total number of negative bags,  $N^+_b$ is the total number of instances in positive bag $b$, and $N^-_a$ is the total number of instances in negative bag $a$. Each data point/instance is either positive or negative, as indicated by the following notation: $\mathbf{x}_{ai}^-$ is the $i^{th}$ instance in the $a^{th}$ negative bag and $\mathbf{x}_{bj}^+$ is the $j^{th}$ instance in the $b^{th}$ positive bag. The $\mu$ and $\sigma^2$ are the mean and variance of the Gaussian function, respectively. For two-class classifier fusion problems in this paper, the positive class (target) is marked label ``$+1$'' and the negative class (non-target) is marked label ``$0$''. The parameter $\mu$ can be set, in this case, to 1, to  encourage the Choquet integral values of positive instances to be 1 and the Choquet integral values of negative instances to be far from 1. Here, the model parameter vector $\boldsymbol{\theta}$ consists of the variance of the Gaussian $\sigma^2$ and the fuzzy measure $\mathbf{g}$ values used to compute the Choquet integral.

However, this previously proposed noisy-or model was subject to user-defined parameter setting such as the variance $\sigma^2$ \cite{du2016multiple}. The variance parameter controls how sharply the Choquet integral values are pushed to 0 and 1, and thus controls the weighting of the two terms separately. A larger variance parameter allows for more noise in the data by allowing points in negative bags to have higher CI values and positive points to have lower CI values. In this paper, we propose new algorithms that can eliminate the variance parameter yet still able to perform effective classifier fusion and handle uncertain labels. 

Furthermore, the previously proposed noisy-or model is a slow algorithmic approach. The noisy-or model needs to compute the valid interval for each fuzzy measure element at each iteration during optimization and can be quite slow. In this paper, we propose a novel optimization technique that will improve the computation time significantly. In addition, the previous noisy-or model  only addresses two-class classification problems. In this paper, however, we shall address both two-class classification problems and real-valued regression problems given remote sensing data with uncertain training labels.

\subsection{Contributions}
The contributions of this paper are four-fold. First, all the newly proposed algorithms work under the multiple instance assumption and can learn from uncertain and imprecise training labels, which is a novelty in remote sensing data fusion. Second, this paper proposes two novel Multiple Instance Choquet Integral models for classifier fusion, which successfully eliminated the variance parameter of previously proposed noisy-or model. Third, this paper proposes a novel Multiple Instance Choquet Integral Regression algorithm, which extends for regression applications where the predictions are real-valued.  Finally, in order to learn the fuzzy measure to be used with the Choquet integral,  a new optimization scheme is proposed in this paper which allows significantly improved computation time as compared to our previous approaches for MIL fusion.

\section{Proposed Algorithms}
\label{sec:algorithm}
This section describes the proposed two Multiple Instance Choquet Integral (MICI) classifier fusion models and a novel Multiple Instance Choquet Integral Regression (MICIR) algorithm for regression applications.

\subsection{The Multiple Instance Choquet Integral Algorithms for Classifier Fusion}

Two Multiple Instance Choquet Integral models, the  min-max and generalized-mean models, are proposed for classifier fusion. The proposed models can perform multi-sensor classifier fusion while learning from uncertain and imprecise training labels. A monotonic normalized fuzzy measure is learned to be used with the Choquet integral to perform two-classs classifier fusion given bag-level training labels. An optimization scheme based on an evolutionary algorithm is used to optimize the models proposed.

\subsubsection{Min-Max Model}
\label{sec:miciminmax}
The min-max model uses ``min'' and ``max'' operators to follow the MIL assumption. The MIL framework assumes that for negative bags, all the instances in the bag are negative (label ``0''). Thus, we write the objective function for the negative bags as: 
\begin{equation}
J_M^- = \sum_{a=1}^{B^-} \max_{\forall \mathbf{x}_{ai}^-  \in \mathscr{B}_a^-} \left(C_\mathbf{g}(\mathbf{x}_{ai}^- ) - 0 \right)^2;  
\label{eq:minmaxobjneg}
\end{equation}

For positive bags, at least one instance in the bag should be positive (label ``1''), so we write the objective function for the positive bags as:
\begin{equation}
J_M^+ = \sum_{b=1}^{B^+} \min_{\forall \mathbf{x}_{bj}^+  \in \mathscr{B}_b^+} \left( C_\mathbf{g}(\mathbf{x}_{bj}^+)-1\right)^2, 
\label{eq:minmaxobjpos}
\end{equation}
where $B^+$ is the total number of positive bags, $B^-$ is the total number of negative bags, $\mathbf{x}_{ai}^-$ is the $i^{th}$ instance in the $a^{th}$ negative bag and $\mathbf{x}_{bj}^+$ is the $j^{th}$ instance in the $b^{th}$ positive bag. $C_\mathbf{g}$ is the Choquet integral output given measure $\mathbf{g}$ computed using (\ref{eq:cg}), $ \mathscr{B}_a^-$ is the $a^{th}$ negative bag, and $\mathscr{B}_b^+$ is the $b^{th}$ positive bag. 

Thus, the objective function for the MICI min-max model classifier fusion approach is written as follows:
\begin{equation}
J_M = J_M^- + J_M^+.
\label{eq:minmaxobj}
\end{equation}

By minimizing the objective function in (\ref{eq:minmaxobj}), we encourage the Choquet integral of all the instances in the negative bag to zero (``$J_M^-$'' term) and encourage the Choquet integral of at least one of the points in the positive bag to one (``$J_M^+$'' term), which fits the MIL assumption. This objective function relies entirely on the training data and labels and does not require user-set variance parameters as needed previously in the noisy-or model. 

\subsubsection{Generalized Mean Model}
\label{sec:micisoftmax}
The ``min'' and ``max'' operators proposed in the above min-max model strictly follows the MIL assumption where only one point is selected for each of the positive bags. Here, in this second model, a generalized mean model is proposed to allow  more points to contribute to the class labels. 

If $p$ is a non-zero real number, and $x_1$,...,$x_n$ are positive real numbers, then the generalized mean with the exponent $p$ of $x_1$,...,$x_n$ is defined as \cite{bullen2003handbook}:
\begin{equation}
M_p(x_1,x_2,...,x_n) = \left( \frac{1}{n} \sum_{i=1}^{n} x_i^p \right)^\frac{1}{p}.
\end{equation}

The generalized mean has the following two properties:
\begin{equation}
\begin{aligned}
M_{-\infty}(x_1,x_2,...,x_n) &= \lim_{p\rightarrow -\infty} M_p(x_1,x_2,...,x_n) \\
&= min(x_1,x_2,...,x_n).
\label{eq:mneginf}
\end{aligned}
\end{equation}
\begin{equation}
\begin{aligned}
M_{\infty}(x_1,x_2,...,x_n) &= \lim_{p\rightarrow \infty} M_p(x_1,x_2,...,x_n) \\
&= max(x_1,x_2,...,x_n).
\label{eq:mposinf}
\end{aligned}
\end{equation}
Therefore, by adjusting the $p$ value, the term can act as varying aggregating operators.

For negative bags, all the instances in the bag are negative (label ``0'') This assumption can be expressed using the generalized mean model as: 
\begin{equation}
J_G^- = \sum_{a=1}^{B^-} \left[ \frac{1}{N_a^-} \sum_{i=1}^{N_a^-} \left(C_\mathbf{g}(\mathbf{x}_{ai}^- ) - 0 \right)^{2p_1} \right]^\frac{1}{p_1} .  
\label{eq:softmaxobjneg}
\end{equation}

Similarly, for positive bags, at least one instances in the bag should be positive (label ``1''): 
\begin{equation}
J_G^+ = \sum_{b=1}^{B^+} \left[ \frac{1}{N_b^+} \sum_{j=1}^{N_b^+} \left( C_\mathbf{g}(\mathbf{x}_{bj}^+)-1\right)^{2p_2} \right]^\frac{1}{p_2} , 
\label{eq:softmaxobjpos}
\end{equation}
where $B^+$ is the total number of positive bags, $B^-$ is the total number of negative bags,  $N^+_b$ is the total number of instances in positive bag $b$, and $N^-_a$ is the total number of instances in negative bag $a$, $\mathbf{x}_{ai}^-$ is the $i^{th}$ instance in the $a^{th}$ negative bag and $\mathbf{x}_{bj}^+$ is the $j^{th}$ instance in the $b^{th}$ positive bag. $C_\mathbf{g}$ is the Choquet integral given measure $\mathbf{g}$, $ \mathscr{B}_a^-$ is the $a^{th}$ negative bag, and $\mathscr{B}_b^+$ is the $b^{th}$ positive bag.  $p_1$ and $p_2$ are the two exponent parameters of the generalized mean functions.

Thus, the objective function for the proposed MICI generalized-mean classifier fusion is written as follows:
\begin{equation}
J_G = J_G^- + J_G^+.
\label{eq:softmaxobj}
\end{equation}

By minimizing the objective function in (\ref{eq:softmaxobj}), we encourage the Choquet integral of all the points in the the negative bag to zero (``$J_G^-$'' term) and encourage the Choquet integral of at least one of the points in the positive bag to one (``$J_G^+$'' term).  The $p$ term allows more points to contribute to the class labels. When $p_1 \rightarrow \infty$ and $p_2 \rightarrow -\infty$, according to properties (\ref{eq:mneginf}) and (\ref{eq:mposinf}), the generalized mean terms becomes equivalent to the min and max operators, making the generalized mean model equivalent to the min-max model. By adjusting the $p$ value, the generalized mean term can act as varying other aggregating operators, such as arithmetic mean ($p=1$) or 	quadratic mean ($p=2$). For another interpretation, when $p \geq 1$, the generalized mean can be rewritten as the $l_p$ norm \cite{rolewicz2013functional}.

\subsection{The Multiple Instance Choquet Integral Regression Algorithm}
A new Multiple Instance Choquet Integral Regression (MICIR) model is proposed to solve regression problems where the desired prediction values are real-valued.  The proposed MICIR algorithm adopts the ``primary-instance'' assumption that there is one primary instance responsible for the label for each bag \cite{ray2001multiple}. The MICIR algorithm fuses multiple sources with real-valued label as well as taking into account the uncertainties in the label.

In parallel with the proposed classifier fusion algorithm, given a set of training data and (real-valued) bag-level training labels, the proposed MICIR learns a fuzzy measure to be used with the Choquet integral and the CI value is used to perform regression on the testing data. The following objective function is proposed to optimize the fitness for regression given a measure $\mathbf{g}$:
\begin{equation}
\min \sum_{b=1}^{N_b} \min_{\forall i, x_{bi} \in X_b} \left( C_\mathbf{g}(\mathbf{x}_{bi}) -d_b\right)^2,
\label{eq:regobj}
\end{equation}
where $N_b$ is the total number of training bags, $d_b$ is the desired training bag-level label for bag $b$, $X_b$ is the set of all the instances in bag $b$, $ C_\mathbf{g}(\mathbf{x}_{bi}) $ is the choquet integral output for the $i^{th}$ instance in bag $b$ given measure $\mathbf{g}$. The Choquet integral output is computed based on Equation (\ref{eq:cg}). The objective function encourages the Choquet integral value of one instance in the bag $b$ to the desired real-valued label $d_b$.

The proposed MICIR algorithm is applicable to  classification problems as well as regression problems with uncertain labels. For regression problems, the desired label $d_b$ is real-valued. The objective function selects the CI value of one instance in the bag $b$ to match the desired real-valued label $d_b$. For two-class classification problems, we set the desired training bag label $d_b=1$ for positive bags and $d_b=0$ for negative bags. In this case, the training data bags need to be re-constructed. The MIL assumes that a bag is labeled negative if all of the instances in the bag is negative. Therefore, each negative instance needs to form its own negative bag. The positive bag can stay the same, as the MIL assumes that bags are labeled positive if at least one instance in the bag is positive. In this case, The objective function encourages all the negative points to label ``0'' and encourages one of the instances in the positive bags to be ``+1''.

% as well by simply regarding the ``0'' and ``+1'' class labels as the desired real-valued predictions.

\section{Optimization}
\label{sec:optimization}
The proposed MICI min-max and generalized-mean models  and the proposed MICIR algorithm learn a fuzzy measure to be used within a Choquet integral for two-class classifier fusion and regression.  The computed Choquet integral with the learned measure is used as the estimated class labels. The goal is to learn the unknown measure from training data and known bag-level training labels and match the computed CI value as closely as possible to the desired labels. To learn the measure, a new optimization scheme based on the evolutionary algorithm is used to optimize the models proposed. 

In previous studies, the valid interval was used to sample measure element values \cite{du2016multiple}. The term ``valid interval'' was define how large the element value can change without sacrificing monotonicity.  The valid interval width for each measure element is set as the difference between the lower and upper bound for each measure element. The lower and upper bounds of a measure element are computed as the largest value of its subsets and the smallest value of its supersets, respectively. Essentially, the valid interval represents how much ``wiggle room'' each measure element has. The previous optimization scheme evaluates the valid intervals of each measure element and samples new element values according to the valid intervals. This method was proven effective in \cite{du2016multiple}, yet it requires a long computation time as the valid interval changes for each measure element after each iteration, and the algorithm needs to evaluate all valid intervals again for the next sample. 

In this section, a new optimization scheme based on the usage of measure element is proposed. First, for all the instances in the training bags, it is easy to obtain which measure element is used for which instance by sorting values from all sources (according to the definition of Choquet integral). Then, a new measure element is sampled according to a multinomial distribution that is based on the counts of how many times a measure element was used in all the training instances. 

Pseudocode for the proposed optimization scheme for both training and testing stages for the proposed models can be seen in Algorithm~\ref{alg:mici}. In the pseudocode, $\mathbf{F}_P^0$ is the fitness values for all measures in the inital population,  $I$ is the maximum number of iterations, $P$ is the measure population size, $\mathbf{\mathscr{G}}$ refers to all measures in the current measure population, $\mathbf{\mathscr{G}}\{p\}$ is the $p^{th}$ measure in measure population $\mathbf{\mathscr{G}}$, $\eta$ is the rate of small-scale mutation,  $\mathbf{F}_P^{t}$ is the fitness values for all measures in Iteration $t$, $F_d$ is the difference of best fitness values between the current and last iteration, and $F_T$ is the threshold of the difference of best fitness value. If $F_d \leq F_T$, the stopping criteria is met and the optimization process stops. The value $F^*$ is the  best (minimum) current fitness value, and $\mathbf{g}^*$ is the best current measure with the highest fitness value.

\begin{algorithm}
\caption{Proposed Optimization Scheme} 
\label{alg:mici}
\begin{algorithmic}[1] 
\Statex  \textbf{TRAINING}
\Require Training Data, Training Labels, Parameters 
\State Initialize a population of measures. Set $t=0$. %\tikzmark{right}\Comment{\ref{sec:measureinit}}   
\State $F^* = max(\mathbf{F}_P^0)$, $\mathbf{g}^* = \argmaxA_{\mathbf{\mathscr{G}}}\mathbf{F}_P^0 $.
\State Evaluate counts of usage of each measure element from training data. %\tikzmark{right}\Comment{\ref{sec:interval}}   
\While { $t < I$ }   
		\For{ $p:=1 \to P$ }
			\State Randomly generate $z \in [0, 1]$.
			\If {$z < \eta $}  % \tikzmark{right}\Comment{\ref{sec:mutation}}   
				\State Sample  $\mathbf{\mathscr{G}}\{p\}$ by small-scale mutation.
			\Else
				\State  Sample  $\mathbf{\mathscr{G}}\{p\}$ by large-scale mutation. 
			\EndIf
		\EndFor
	\State  Evaluate fitness of sampled measures using (\ref{eq:minmaxobj}) or (\ref{eq:softmaxobj}) or (\ref{eq:regobj}), depending on classification or regression problems.
	\State  Select measures.  %\tikzmark{right}\Comment{\ref{sec:selection}} 
		\State Compute $ F_d = |max(\mathbf{F}_P^{t}) - F^*|$.
	\If {$ max(\mathbf{F}_P^{t}) > F^*$}
					\State $F^* = max(\mathbf{F}_P^{t})$, $\mathbf{g}^* = \argmaxA_{\mathbf{\mathscr{G}}}\mathbf{F}_P^{t} $.
	\EndIf
	\If {$F_d \leq F_T$}
	break;
	\EndIf
	\State $t \gets t+1$.
\EndWhile

\Return $\mathbf{g}^* $
\Statex 
\Statex  \textbf{TESTING}
\Require Testing Data, $\mathbf{g}^* $
\State $TestLabels \gets$ Choquet integral output computed based on Equation (\ref{eq:cg}) using the learned $\mathbf{g}^*$ above.

\Return $TestLabels$
\end{algorithmic}
\end{algorithm}

\subsection{Measure Initialization}
\label{sec:measureinit}
In the algorithm, a population (size $P$) of the Choquet integral measures is generated and each measure in the population is intialized randomly to a set of values between $[0, 1]$ that satisfies monotonicity. 

Two types of initialization approaches, ``top-down'' and ``bottom-up'' approaches, were implemented. In the ``top-down'' initialization, the values of the measure elements were sampled from the top of the lattice towards the bottom. Suppose we have four sources to fuse ($m=4$), for example, the $(m-1)$-tuple measure elements ($g_{123}$, $g_{124}$, $g_{134}$, and $g_{234}$) were first sampled randomly between 0 and 1. Then, the $(m-2)$-tuple measure elements ($g_{12}$, $g_{13}$, $g_{14}$, $g_{23}$, $g_{24}$, and $g_{34}$) were sampled between 0 and its corresponding superset. For example, $g_{12}$ measure element value is sampled randomly between 0 and $\min(g_{123}, g_{124})$, due to the monotonicity property of fuzzy measures. The process goes on until the singletons ($g_1, g_2, g_3, g_4$) were each sampled between 0 and their corresponding superset values.

Similarly, the ``bottom-up'' approaches samples measure element values from the bottom of the lattice up. First the singletons were sampled between 0 and 1 randomly. Then, the duples were sampled between its corresponding subsets and 1. For example, in the four-source case, the initial value of $g_{12}$ was sampled randomly between $\max(g_1, g_2)$ and 1. The process goes on until the $(m-1)$-tuple measure elements were sampled, thus initializing all the element values in the entire measure. Note that the measure element corresponding to the full set is always equal to 1 (e.g. $g_{1234} \equiv 1$ for four sources), according to the normalization property of the fuzzy measure.

In our experiments, the two initialization approaches yield different sets of measure element values but seem to have little impact on the final measure learned. In the following experiments, the measure is initialized by randomly flipping a coin and pick either ``top-down'' or ``bottom-up'' initialization approach. 

\subsection{Evaluation of Counts of Usage of Measure Elements}
\label{sec:interval}
The counts of usage of measure elements can be obtained directly from training data by sorting values from source values in training. For example, three sensors yield 0.8, 0.2, 0.1 values, respectively, for a data point. To fuse these results using the CI, according to the definition of Choquet integral, $g_1$, $g_{12}$, and $g_{123}$ will be used. This process is repeated for all training data points and the count of how many times each measure element will be used in the CI is recorded. The count value will be recorded as a vector $\mathbf{v}=\{v_1, v_2, ... v_{2^m-1}\}$ for $m$ fusion sources. The count value vector has the same length as the fuzzy measure vector.

\subsection{Mutation}
\label{sec:mutation}
In the optimization, mutations of two different scales were designed in search of the optimal solution. The measure element usage count obtained above are used in determining which measure element to sample during both types of mutations.

In the small-scale mutation, only one measure element is sampled. The element to be sampled is chosen by randomly sampling from a multinomial distribution based on the counts of how many times a measure element was used in all the training instances. The probability of sampling a particular measure element $g_l$ is set to
\begin{equation}
P(g_l) = \frac{v_l}{\sum_{o=1}^{2^m-1}v_o},
\label{eq:pl}
\end{equation}
where $v_l$ is the number of times measure element $g_l$ is used in training data. The measure element that was used most frequently by the training data to compute the Choquet integral will have the largest probability to be updated. In the large-scale mutation, all the measure elements are sorted in descending order based on the number of times it was used by the training data and all measure elements are updated according to the sort order. The new measure values are sampled from a truncated Gaussian (TG) distribution. The details of truncated Gaussian sampling method can be seen in \cite{du2017multiple}.

The rate of the small-scale mutation $\eta \in [0,1]$ is defined by users. The rate of large-scale mutation is $1-\eta$.

\subsection{Selection}
\label{sec:selection}
The measures retained for the next generation are selected based on their fitness function values computed using Equation (\ref{eq:minmaxobj}) or (\ref{eq:softmaxobj}) or (\ref{eq:regobj}), depending on using the min-max model, generalized-mean model, or the MICIR algorithm for classifier fusion or regression. In each iteration, all measures in the population are sampled, yielding a child measure population of size $P$. The measure population before sampling is regarded as the parent measure population (size $P$). Both the parent and child measure populations are pooled together (size $2P$) and their fitness values are computed using Equation (\ref{eq:minmaxobj}), (\ref{eq:softmaxobj}), or (\ref{eq:regobj}). Then, $P/2$ measures with the top 25\% fitness values are kept and carried over to the next iteration (elitism), and the remaining $P/2$ measures to be carried over are sampled according to a multinomial distribution based on their fitness values from the remaining 75\% of the parent and child population pool, following a similar approach to Equation \eqref{eq:pl}. Among the new measure population, the measure with the highest fitness value is kept as the current best measure $\mathbf{g}^*$. The process continues until a stopping criterion is reached, such as when the maximum number of iterations is reached or the change in the objective function value from one iteration to the next is smaller than a fixed threshold. 

At the end of training process, the best measure $\mathbf{g}^*$ with the best fitness value so far is returned as the learned measure and used for testing. Note that we are minimizing the objective function, so the minimum fitness value is regarded as the best fitness value.

\section{Experimental Results}
\label{sec:experiments}

This section describes experiments and presents results of the proposed algorithms on synthetic as well as real remote sensing data. Synthetic data sets were generated to evaluate the effect of parameters in both classification and regression. In addition, the MUUFL Gulfport HSI data set and the HARVIST crop yield data set was used for real target detection and crop yield prediction applications.

\subsection{Synthetic Classification Data Set}
\label{sec:classsyn5source}

A synthetic 5-Source classification data set was constructed to investigate the effect of ``contamination'' in the training bags. ``Contamination'' is defined as there are positive points mixed in the negative bags. The data set was constructed with 100 bags and 10 data points per bag. Half of the bags are positive (with label ``1'') and half are negative (with label ``0''). In the following experiments,   $P=30$, $\sigma^2=0.1$, $I=5000$, $F_T=0.0001$, $\eta=0.8$, $\mu=1$, $p_1=10$, and $p_2 = -10$.  %According to the assumption of MIL, the negative bags must contain all negative points but the positive bag only needs to contain at least one positive points. 

Assume half of the training bags are positive and contain 100\% positive points. Then, a varying percentage of positive points were added to the negative bags so that the negative bags were ``contaminated'' in various degrees.  Relative error \cite{wagstaff2008multiple} is used to evaluate the performance for the MICI classification models:
\begin{equation}
Error_{reg}(y,\hat{y}) = \left | y-\hat{y}\right|,
\end{equation}
where $y$ is the true label (``1'' or ``0'') and $\hat{y}$ is the estimated label for each data point.

The relative error results (across all data points) with MICI noisy-or model, min-max model, and generalized mean model were presented in Table~\ref{table:ratiocontaminationsclNO}.  As can be seen, in all three models, the relative error generally increases as the percentage of contamination increases, which is as expected. When the percentage of contamination goes towards 100\%, the error is nearly 1 (100\% wrong) as well, which makes sense as the negative bags are now filled with positive points and it is not possible to distinguish positive and negative points. Note that the proposed min-max model optimizes only one instance in each bag based on the standard MIL assumption. In the min-max model, depending on the random generalization of this synthetic data set, the remaining instances (other than the max and min instance) in the bags can have varying CI values and will impact the overall relative error.

\begin{table}[h]
\centering
\caption{Relative error versus contamination (denoted as ``con.'', in percentage) for synthetic classification data set for MICI noisy-or model, min-max model, and generalized-mean model across five runs. The standard deviation is noted in parentheses.}
\label{table:ratiocontaminationsclNO}
\resizebox{0.48\textwidth}{!}{
\begin{tabular}{ |c|c|c|c|}
  \hline                  
 \textbf{ Con. (\%)  }   & \textbf{ noisy-or} & \textbf{ min-max} & \textbf{ generalized-mean}\\
\hline\hline 
0\% & 0.371(0.026) &0.288(0.025) & 0.247(0.037)\\ \hline
10\% & 0.492(0.008) &0.514(0.003) &0.517(0.002 \\ \hline
20\% & 0.534(0.007) &0.536(0.011) &0.551(0.006) \\ \hline
30\% & 0.588(0.007) &0.570(0.017) &0.595(0.005) \\ \hline
40\% &0.630(0.012) &0.655(0.005) &0.660(0.002) \\ \hline
50\% &0.668(0.004) &0.686(0.005) &0.694(0.007) \\ \hline
60\% &0.697(0.014) &0.255(0.171) &0.709(0.020) \\ \hline
70\% &0.751(0.012) &0.217(0.191) &0.744(0.022) \\ \hline
80\% &0.787(0.019) &0.359(0.304) &0.783(0.027) \\ \hline
90\% &0.812(0.012) &0.579(0.255) &0.821(0.017) \\ \hline
100\% & 0.868(0.014) &0.345(0.366) &0.882(0.011) \\ \hline 
\end{tabular}
}
\end{table}

\subsection{Synthetic Regression Data Set}  
\label{sec:simregdata}
A synthetic 5-source regression data set were constructed to investigate the performance of the proposed MICIR regression algorithm. In this experiment, the percentage of primary instances in the training bags and signal-to-noise ratio (SNR) changes, and the performance of the proposed MICIR algorithm is observed. All data sets used in this section have 1000 data points. Each data point has 5 dimensions. The data points were grouped into 10 bags, each bag with 100 data points. The data points in this regression data set have real-valued labels between $[0, 1]$. Relative error \cite{wagstaff2008multiple} is used to evaluate the performance for the MICIR algorithm:
\begin{equation}
Error_{reg}(y,\hat{y}) =
\begin{cases}
     \left| \frac{y-\hat{y}}{y}\right|,  &  \text{if }  y \in (0,1]\\
     \left| {y-\hat{y}}\right|,          & \text{if }  y=0
\end{cases}
\end{equation}
where $y$ is the true label and $\hat{y}$ is the estimated label for each data point.

The proposed MICIR algorithm operates under the assumption that only one primary instance is associated with the label of each bag \cite{ray2001multiple}. First, we vary the percentage of primary instances in the bags to observe the relationship between the percentage of primary instances and the performance of proposed MICIR algorithm. The percentage of primary instances in the bag takes the  value of 0\% to 100\% with an increment of 10\%. For each bag, the ``primary instances'' have label values that are exactly the same as the bag-level training label. The non-primary instances have randomly generated label values (different from the bag-level labels) that are generated from a completely random measure. Table~\ref{table:ratioposneg} shows the relationship between the percentage of primary instances in the bag and the mean relative error over all the data points across five runs.  As can be seen from the table, when the percentage of primary instances in the bag increases, the relative error decreases.

\begin{table*}[h]
\centering
\caption{Relative error versus percentage of primary instances for synthetic regression data set for MICI Regression model across five runs. The standard deviation is noted in parentheses.}
\label{table:ratioposneg}
\resizebox{\textwidth}{!}{
\begin{tabular}{ |c|c|c|c|c|c|c|}
  \hline  
& \multicolumn{6}{c|}{\textbf{Percentage of primary instances}}\\
    \cline{2-7}                  
        & 0\% & 10\% &20\%& 30\%& 40\% & 50\%\\
\hline
Relative Error   & $0.730(0.001)$    & $0.488(0.001)$   &$0.492(0.000)$      & $0.451(0.001)$     & $0.365(0.000)$      & $0.301(0.001)$  \\
    \cline{1-7}                  
      & 60\% & 70\% &80\%& 90\%& 100\% & \\
\hline
Relative Error  & $0.345(0.000)$  &  $0.274(0.001)$   & $0.104(0.001)$      & $0.059(0.002)$   & $0.002(0.002)$       &     \\
  \hline 
\end{tabular}
}
\end{table*}

Second, the performance of MICIR is observed with varying levels of signal-to-noise ratio (SNR). In this experiment, 100\% of the points in the bag are primary instances and varying amount of noise is add to the entire data set, creating SNR value from 50dB to 0dB with an increment of -5dB. Table~\ref{table:snr} shows the relationship between the SNR values in the bag and the mean relative error over all the data points across five runs.  As can be seen from the table, the relative error decreases when the SNR value increases, as expected.

\begin{table*}[h]
\centering
\caption{Relative error versus SNR for synthetic regression data set MICI Regression model across five runs. The standard deviation is noted in parentheses.}
\label{table:snr}
\resizebox{\textwidth}{!}{
\begin{tabular}{ |c|c|c|c|c|c|c|}
  \hline  
& \multicolumn{6}{c|}{\textbf{SNR value}}\\
    \cline{2-7}                  
        & 0dB & 5dB &10dB  & 15dB  & 20dB &25dB \\
\hline
Relative Error   & $0.691(0.027)$    & $0.443(0.029)$   &$0.266(0.013)$      & $0.173(0.025)$     & $0.101(0.013)$      & $0.061(0.012)$  \\
    \cline{1-7}                  
      & 30dB & 35dB & 40dB  &45dB  & 50dB & \\
\hline
Relative Error  & $0.044(0.003)$  &  $0.021(0.003)$   & $0.014(0.002)$      & $0.008(0.002)$   & $0.005(0.001)$       &     \\
  \hline 
\end{tabular}
}
\end{table*}

\subsection{MUUFL Gulfport Target Detection Experiments}
\label{sec:targetdectionmici}
The proposed MICI min-max model and generalized-mean model were tested on a real target detection application using the MUUFL Gulfport hyperspectral data set. The MUUFL Gulfport hyperspectral data set \cite{gader2013muufl} was collected over the University of Southern Mississippi Gulf Park Campus. The data set used in this experiment consists of three hyperspectral data cubes collected on three separate flights at an altitude of 3500' over the campus area.  The HSI data cubes have a ground sample distance of 1m.\footnote{The data set is available at https://github.com/GatorSense. The three flights used in this experiment corresponds to ``muufl\_gulfport\_campus\_w\_lidar\_1.mat'', ``muufl\_gulfport\_campus\_3.mat'', and ``muufl\_gulfport\_campus\_4.mat''. }  The image from campus 1 is $325\times337$ pixels in size. The image from campus 3 is $329\times345$ pixels in size.  The image from campus 4 is $333\times345$ pixels in size. All HSI data cubes contain 72 bands corresponding to wavelengths 367.7nm to 1043.4nm and were collected using the CASI hyperspectral camera \cite{gader2013muufl,jiao2015functions}. In this experiment, the first four and last four noisy bands were removed.

\begin{figure}
\centering
\begin{subfigure}[t]{0.99\linewidth}
\centering
\includegraphics[width=\textwidth,trim={10mm 7mm 10mm 10mm},clip]{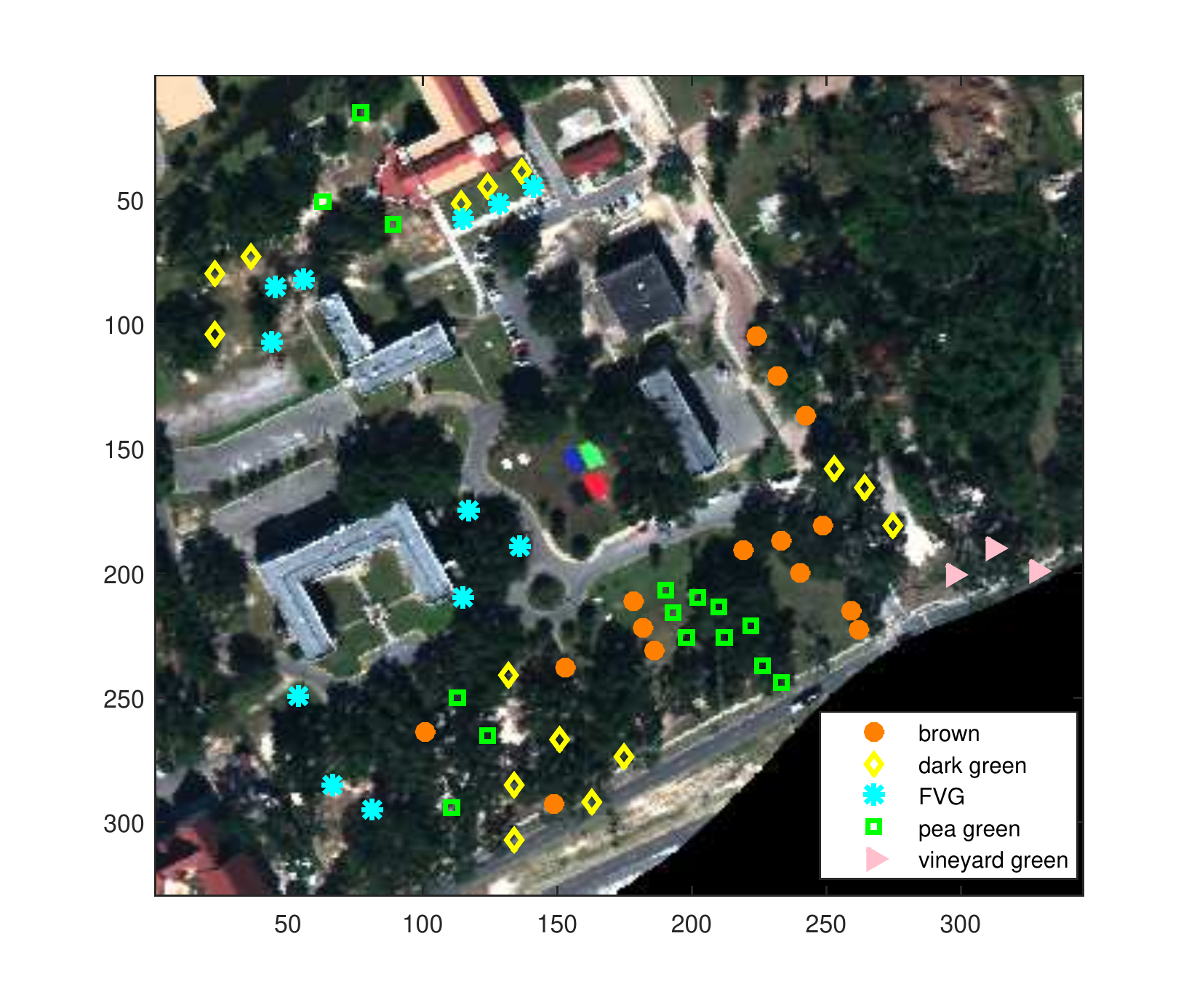}
\end{subfigure}
\caption[The RGB image from MUUFL Gulfport ``campus 3'' data set. ]{The RGB image from MUUFL Gulfport ``campus 3'' data set. The colored markers show the true target locations for all target types.}
\label{fig:flights}
\end{figure}

A total of sixty cloth panel targets were placed in the scene.  The targets were cloth panels of five different colors: fifteen brown, fifteen dark green, twelve faux vineyard green (FVG), fifteen pea green, and three vineyard green (vineyard green targets are not considered in this experiment as the target signature was not available and target number too small). The goal is to find all (brown, dark green, FVG, and pea green)  targets in the scene. These targets varied from sub-pixel targets (at $0.25m^2$ corresponding to a quarter of a pixel in area) up to super-pixel targets (at $9m^2$) with varying levels of occlusion. For each target, a GPS ground truth location was collected using a Trimble Juno SB hand-held device.  Figure~\ref{fig:flights} shows the RGB images of the two flights over the campus and the GPS target locations.  In this experiment, the GPS device has accuracy up to 5m.  Therefore, the groundtruth locations for each target are only accurate within a $5\times5$ pixel halo.  The MIL approach fits the problem well and the proposed MICI min-max and generalized-mean models were used to perform target detection given the inaccuracy in the groundtruth labels.

The adaptive coherence estimator (ACE) detector \cite{scharf1996adaptive,kraut2005adaptive,pulsone2000computationally} was applied to the imagery using spectral signatures of four of the target types (as spectral signatures for these targets were available from previous studies\footnote{The target spectra used in this experiment come from ``tgt\_img\_spectra .mat'' in the data set. The four target types are: brown, dark green, FVG and pea green. }).  The background mean and background covariance for the ACE detector was estimated using the global mean and covariance of all the pixels in each image. The ACE results used in this experiment were normalized between $[0, 1]$.  %Note that the signed ACE detector used yields confidence values between $[-1, 1]$. Therefore, all ACE results used in this experiment were normalized to be between $[0, 1]$. %zero and one by adding 1 to the original ACE results and divide by 2. 

Cross validation is performed on this data set, i.e., training on campus 1 and testing on campus 3 and campus 4, and so on. First, the mean and covariance of the training imagery, $\mu_{tr}$ and $cov_{tr}$, were computed. Then, the ACE detection map for the training imagery, $ACE_{tr}$, was obtained using $\mu_{tr}$ and $cov_{tr}$. Each pixel in the detection map has four dimensions, each dimension corresponds to ACE confidence values for four target types.  Each of the ACE results highlights different locations corresponding to different targets. The four ACE confidence maps are the sources to be fused by the proposed MICI models.

To construct training bags, a $5\times5$ window was put around each groundtruth target location of the training imagery. Each window forms a positive bag and all the pixels in the windows are instances in the positive bag.  The size of the positive bag corresponds to the accuracy of the GPS device used to collect the groundtruth points.  One negative bag was constructed by randomly picking 1600 background pixels that do not belong to any of the windows. The positive bags were labeled ``1'' and the negative bag was labeled ``0''. The proposed models were applied to the training data and a fuzzy measure $\mathbf{g^*}$ is learned.

In testing, the ACE detection results for the test imagery, $ACE_{te}$, were computed using training mean and covariance $\mu_{tr}$ and $cov_{tr}$. Then, the Choquet integral given the learned measure  $\mathbf{g^*}$ and $ACE_{te}$ was computed as the test fusion result.

%Experiments were conducted on the (un-normalized) HSI data.

\begin{table*}[h]
\centering
\caption{The AUC results at on un-normalized MUUFL Gulfport data across five runs. The standard deviation is noted in parentheses.}
\normalsize{
\resizebox{\textwidth}{!}{
\begin{tabular}{p{2cm}|c|cccccc}
\hline
\textit{Notes} & Methods & \textbf{Train1Test3}  & \textbf{Train1Test4}  & \textbf{Train3Test1}  & \textbf{Train3Test4} & \textbf{Train4Test1} & \textbf{Train4Test3} \\
  \hline \hline
\multirow{4}{\hsize}{\textit{Sources: individual target types.}} &  Brown  & $0.265$  & $0.264$  & $0.334$  &  $0.267$ & $0.307$ &  $0.263$  \\ 
 & Dark Green & $0.266$  & $0.261$  & $0.328$  &  $0.256$ & $0.293$ &  $0.266$  \\ 
 & FVG & $0.114$  & $0.106$  & $0.122$  &  $0.107$ & $0.136$ &  $0.109$  \\ 
 & Pea Green & $0.088$  & $0.000$  & $0.107$  &  $0.000$ & $0.100$ &  $0.091$  \\ \hline 
\multirow{5}{\hsize}{\textit{Instance- based fusion methods in comparison.}} & SVM  & $0.185$  & $0.195$  & $0.164$  &  $0.175$ & $0.245$ &  $0.220$  \\
 & min  & $0.000$  & $0.000$  & $0.046$  &  $0.073$ & $0.026$ &  $0.023$  \\
 & max  & $\underline{0.345}$  & $0.329$  & ${0.459}$  &  $\underline{0.339}$ & $\mathbf{0.349}$ &  $0.328$  \\
 & mean  & $0.224$  & $0.221$  & $0.269$  &  $0.214$ & $0.260$ &  $0.246$  \\
 & CI-QP   & $0.328$  & $0.325$  & $0.399$  &  $0.330$ & $0.260$ &  $0.272$  \\ \hline 
\multirow{3}{\hsize}{\textit{MIL comparison methods.}} & mi-SVM & $\mathbf{0.346}$  & $\mathbf{0.337}$  & $0.350$  &  $0.293$ & $0.317$ &  $0.317$  \\ 
 & EM-DD & $0.062(0.014)$  & $0.073(0.013)$  & $0.002(0.003)$  &  $0.021(0.004)$ & $0.005(0.008)$ &  $0.021(0.021)$  \\ 
  & MICI Noisy-Or  & $\mathbf{0.346(0.000)}$ & $\underline{0.331(0.001)}$ &$0.461(0.000)$  & $\mathbf{0.340(0.000)}$ &$\mathbf{0.349(0.000)}$  & $\underline{0.329(0.000)}$ \\ \hline 
\multirow{2}{\hsize}{\textit{Proposed MICI models.}} 
 & MICI Min-Max  & $\underline{0.345(0.001)}$ & $\underline{0.331(0.008)}$ &$\underline{0.463(0.010)}$  & $0.333(0.013)$ &$\underline{0.348(0.002)}$  & $\underline{0.329(0.001)}$ \\ 
 & MICI Generalized Mean & $\underline{0.345(0.000)}$ & $\underline{0.331(0.005)}$ &$\mathbf{0.466(0.004)}$  & $\underline{0.339(0.001)}$ &$\mathbf{0.349(0.001)}$  & $\mathbf{0.330(0.002)}$ \\ 
\hline
\end{tabular}}}
\label{table:muuflunnormAUC}
\end{table*}

The proposed MICI min-max and MICI generalized-mean models were first compared with the four ACE results using the four individual target signatures  (Brown, Dark Green, FVG, and Pea Green). The fusion results were then compared with a standard Support Vector Machine (SVM) on the four ACE maps, and taking the max, min, or mean over the four ACE results. The proposed models were also compared with CI-QP \cite{grabisch1996application}, mi-SVM \cite{andrews2002support}, EM-DD \cite{zhang2001dd}, and the previously proposed MICI noisy-or methods. The CI-QP approach learns a fuzzy measure for Choquet integral by optimizing a least squares error objective using Quadratic Programming.  The CI-QP approach assumes an accurate label for every training data point and, thus, does not inherently support MIL-type learning.  In our application of CI-QP to this problem, we gave all points in a positive bag the label of ``1'' and all points in the negative bag as a label of ``0''. The mi-SVM and EM-DD are both widely used multiple instance learning approaches as discussed in the literature review. % The mi-SVM method is a Support Vector Machine classification approach with an extension for Multiple Instance Learning. The EM-DD is an Expectation-Maximization (EM) version of the diverse density algorithm \cite{dietterich1997solving}, which iteratively selects an instance from the training bags and updates the estimated target ``concept'' based on the selected instances. 
%All the comparison methods were run on un-normalized MUUFL Gulfport data.
 
The receiver operating characteristic (ROC) curve is used to evaluate the target detection results. The ROC curve plots the positive detection rate (PD, Y-axis) against the false alarm rate (FAR, X-axis). The performance of the algorithms was evaluated by computing the area under curve (AUC) results with FAR up to $1\times 10^{-3}/m^2$ (corresponding to a reasonable scale of 1 false alarm in 1000$m^2$). % and (2) the positive detection rate versus the false alarm rate.

Table~\ref{table:muuflunnormAUC}  shows the AUC results at FAR up to $1\times 10^{-3}/m^2$ for all fusion methods with un-normalized hyperspectral data. In the AUC table, the ``best'' performance was determined by comparing the mean value of the AUC results, and in the case where the mean is the same between methods, the one result with smaller standard deviation is preferred.  The best two results were \textbf{bolded} and \underline{underlined}, respectively. It can be seen from Table~\ref{table:muuflunnormAUC} that the proposed  MICI min-max and MICI generalized-mean models have mostly bolded and/or underlined (best and second best) results. % Table~\ref{table:muuflnormAUC}, Table~\ref{table:muuflnorm01AUC}, and Table~\ref{table:muuflnormdistAUC}   show the AUC results at FAR up to $1\times 10^{-3}/m^2$ for all fusion methods with normalized hyperspectral data using the three normalization methods as shown in Equations (\ref{eq:norm1})-(\ref{eq:norm3}), respectively. 

\begin{figure}
\centering
\begin{subfigure}[t]{0.99\linewidth}
\centering
\includegraphics[width=\textwidth,trim={0mm 7mm 10mm 8mm},clip]{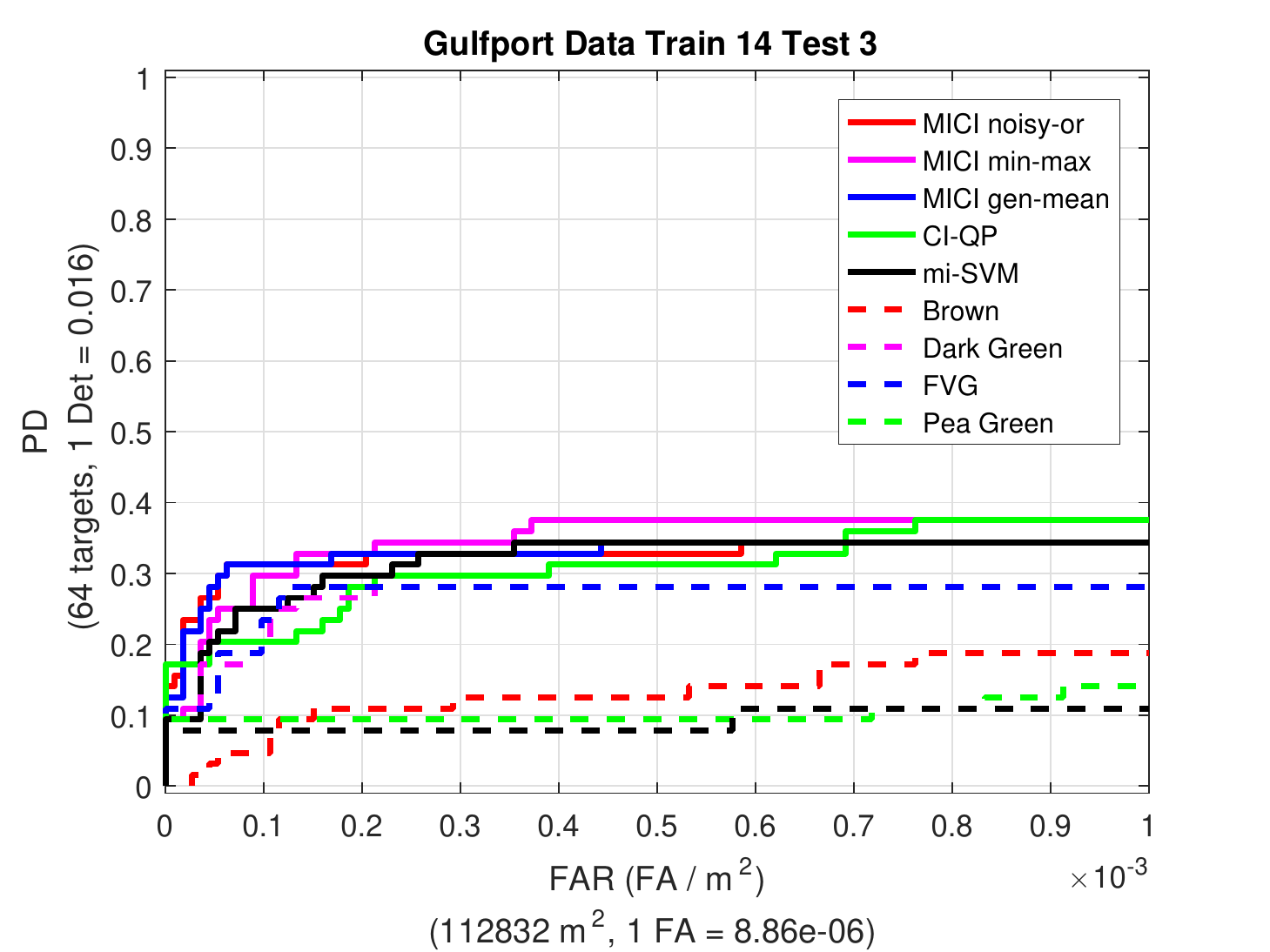}
\end{subfigure}
\caption{The ROC curve for one run of train on campuses 1 and 4, test on campus 3.}
\label{fig:gulfport_train14test3_run1}
\end{figure}

Figure~\ref{fig:gulfport_train14test3_run1} shows the ROC curve plot after one run using unnormalized campuses 1 and 4 data as training data and test on campus 3 data. Figure~\ref{fig:gulfport_train14test3_run1} provides a visual ROC curve example to complement the AUC table results. The remaining cross validation experiments yield similar ROC curve results. The X-axis of the ROC curves represents the False Alarm Rate (FAR) between $[0, 0.001]$ and the Y-axis represents Positive Detection (PD). 

As can be seen from Table~\ref{table:muuflunnormAUC} and Figure~\ref{fig:gulfport_train14test3_run1},  detection results of individual sources are relatively poor, which is understandable as the target types and signature vary greatly and it is preferable to fuse detector outputs for all target types rather than use individual target detector. The CI-QP method (which uses  CI fusion but does not work with uncertain labels) and the mi-SVM method (which works under MIL but does not use CI for fusion) are both outperformed by the proposed MICI methods in general (which uses both MIL and CI techniques). Both proposed MICI models, the min-max and generalized-mean models, yield satisfactory results on target detection. The generalized-mean model has slightly higher overall AUC and higher detection rate compared with the min-max model in the lower FAR range. This is due to the fact that the min-max model only considers the minimum and maximum instance in the bag, while the generalized-mean model has a more relaxed constraint. In the case where there are more than one target pixels in a positive bag (which can be the case in the MUUFL Gulfport data set), the min-max model will likely only encourages one instance to be positive, while the generalized-mean model can assess more instances.

\begin{table*}[h]
\centering
\caption{RMSE error for CA and KS corn and wheat yield. The two results with the lowest errors were \textbf{bolded} and \underline{underlined}, respectively. The unit is bushels per acre. The standard deviation across three runs is noted in parentheses.}
\label{table:testerror2005CA}
  \begin{tabular}{ |p{3cm}|c|c|c|c|c|}
    \hline
\textit{\scriptsize{Notes}} & \textbf{Regression Methods} & \textbf{Wheat CA} & \textbf{Corn CA} & \textbf{Wheat KS} &\textbf{Corn KS}\\
\hline \hline
\multirow{3}{\hsize}{\textit{\scriptsize{Produces instance-level labels. Fusion sources.}}} & Linear Regress & $8.32(0.00)$ & $ 3.17( 0.00) $ & $ \underline{2.08( 0.00)} $ &  $ 16.28 ( 0.00)$ \\    %\cline{2-4}
& RVM & $\underline{ 3.20 ( 0.00)} $ &  $\mathbf{ 0.22( 0.00)}$ & $ 2.15 ( 0.21)$ &  $\mathbf{ 13.48(0.75) }$\\  %\cline{2-4}
& SVR &  $ 8.60( 0.14) $ &  $ 9.38( 0.01) $  & $ 3.35 ( 0.37) $ &  $ 20.15(1.39) $\\
\hline
\textit{\scriptsize{Proposed regression method.}} & MICI Regression Fusion &  $ \mathbf{1.57( 0.27)} $ &  $\underline{ 1.99( 0.00) }$  & $ \mathbf{1.98( 0.00}$ &   $ \underline{15.02( 0.00}$\\\hline %\cline{2-4}
\multirow{4}{\hsize}{\textit{\scriptsize{Fusion methods in comparison.}}} & Another layer of RVM/SVR &  $ 3.30( 2.50) $ & $ 4.14( 1.25) $ & $3.35 ( 1.02) $ &  $26.73 ( 10.54)$  \\%\cline{2-4}
& Taking the max &  $ 11.92( 0.67)$ & $ 12.72( 0.61)$  & $ 2.85(0.16) $ &  $ 34.39(10.83) $    \\%\cline{2-4}
& Taking the min &  $ 4.02( 0.18 )$ &  $ 4.23( 0.06)$  & $ 9.04(1.84) $ &  $ 20.77(7.12) $\\%\cline{2-4}
& Taking the mean &  $ 7.08 ( 0.09)$ &  $4.81 ( 0.12)$ & $3.40 ( 0.19)$ &  $20.71 ( 1.48)$  \\
\hline
\multirow{3}{\hsize}{\textit{\scriptsize{Direct bag-level label prediction.}}}& Cluster MIR & $ 10.17( 0.00)$ &  $11.19 ( 0.00)$  & $ 8.09(0.00)$ &  $ 31.74( 0.00)$ \\
%\cline{2-4}
& Aggregate MIR & $ 10.05 ( 0.00 )$ &  $ 11.19( 0.00)$  & $ 6.87(0.00) $ &  $ 29.97( 0.00)$ \\
%\cline{2-4}
& RFC-MIR & $ 16.57( 0.00 )$ &  $ 15.63(0.00) $  & $6.55 (0.00)$ &  $ 33.20( 0.00)$  \\
\hline
  \end{tabular}
\end{table*}

\subsection{Crop Yield Data Set Experiments}

\begin{table}[h]
\centering
\caption[Number of counties (bags) with both corn and wheat yield in the crop yield data set.]{Number of counties (bags) with both corn and wheat yield in the crop yield data set \cite{wagstaff2008multiple}.}
\label{table:countynum}
  \begin{tabular}{ |c|c|c|c|c|c|}
    \hline
   & \multicolumn{4}{c|}{\textbf{Training}} & \textbf{Testing}\\
\hline
 \textbf{ Year} &\textbf{ 2001} &\textbf{2002} &\textbf{2003 }&\textbf{2004} &\textbf{2005}\\
\hline
\textbf{ CA }& 17 &16 &18 &15 &13\\
\hline
\textbf{ KS} & 100 &102 &100 &102 &98\\
\hline
  \end{tabular}
\end{table}

This section presents results on fusion with real-valued prediction values (regression) using the proposed Multiple Instance Choquet Integral Regression (MICIR) algorithm. Experiments were conducted on a real crop yield prediction application on a MODIS remote sensing data set provided from the HARVIST (Heterogeneous Agricultural Research Via Interactive, Scalable Technology) project \cite{wagstaff2007salience, wagstaff2008multiple,wang2012mixture}.

The crop yield data set contains MODIS data observations of corn and wheat yield in the states of California and Kansas over 5 years (2001-2005). There are 100 randomly selected pixels included for each county. The surface reflectance values were reported for each pixel containing 92 values: observations in red for 46 timepoints (every 8 days across the year) and observations in infrared (IR) for 46 timepoints (every 8 days across the year). The zeros and ``-32767'' reflectance values were indicated as ``bad values'' in the original data set and are removed for this experiment. Only the counties that reported both corn and wheat yield values are considered and Table~\ref{table:countynum} shows the number of counties that are considered in the states of California and Kansas across the years.

This data set suits the multiple-instance framework well as each county can be naturally regarded as a bag with multiple data collections (instances). In this experiment, the corn and wheat yield values for each county are regarded as the real-valued regression labels. As the Choquet integral works with values between zero and one, the corn and wheat yield values were normalized between zero and one by Equation (\ref{eq:yncrop}):
\begin{equation}
Y_n = \frac{Y-Y_{min}}{Y_{max}-Y_{min}},
\label{eq:yncrop}
\end{equation}
where $Y_n$ is the normalized corn or wheat yield value that will be used as the regression training labels. $Y_{min}$ and $Y_{max}$ are the min and max yield value in training, respectively.

Linear regression, Relevant Vector Machine (RVM) regression \cite{tipping2000the, tipping2001sparse}, and Support Vector Regression (SVR) \cite{drucker1997support} were applied to the data. These three regressors operate on all instances and each give  a set of instance-based labels (these are essentially Instance-MIR methods with different regressors). We used Gaussian kernel for both RVM and SVR methods. Then, the MICI Regression model is applied to fuse these three regressors and compared to use another layer of RVM and/or SVR (whichever one gives better performance) or simply taking the max, min, or mean of the regression sources as a fusion method. The results are also compared with existing multiple instance regression methods such as the Aggregate MIR, Cluster MIR, and Robust Fuzzy Clustering MIR (RFC-MIR). 

The test error was computed by computing the root mean squared error (RMSE) between the predicted county-level (bag-level) yield values for the test year and the (known) actual county-level yield values for the test year, as follows:
\begin{equation}
RMSE = \sqrt{\frac{\sum_{b=1}^{B}\left ( \hat{y_b} - y_b \right )^2}{B}},
\label{eq:yncrop}
\end{equation}
where $B$ is the number of (test) bags, $\hat{y_b}$ is the predicted county-level yield values, and $y_b$ is the (known) actual county-level yield values for the test year.

Table~\ref{table:testerror2005CA}  presents the RMSE prediction error results for corn and wheat yield for the states of California (CA) and Kansas (KS), using crop yield training data from years 2001-2004 and testing on year 2005.  From the table, we can first of all see that all methods are able to predict the yields compariable to results from previous literature such as \cite{wang2012mixture}. The proposed MICI Regression method produces the lowest or the second lowest error across both states and both crop types. Naturally, the performance of MICI will depend on the performance of input sources (in this case, three instance MIR approaches linear regression, RVM and SVR). The performance of MICI Regression, from Table~\ref{table:testerror2005CA}, has surpassed each of its sources as well as other fusion such as taking the max and min. RVM performs well especially for corn, but RVM requires the computation of the kernel matrix, which can be hard when the input data set has higher dimensions. RVM is also highly dependent on the choice of kernel and the parameters such as the kernel width.

\section{Discussion on Running Time}
\label{sec:runtime}
%In the proposed algorithms, the most computational consuming step is computing the Choquet integral for each data point to be used with the objective function of the proposed models. The Choquet integral has $O\left(2^m\right)$ parameters for $m$ fusion sources. The computational complexity of $I$ iterations across $B$ bags and $N$ data points is $O\left(INB2^m\right)$. 

Experiments were conducted to compare the computing time of the newly proposed optimization scheme based on the usage of measure elements in the training data and the previously proposed optimization schemes based on valid intervals. The experiment used here is the MUUFL Gulfport target detection experiment same as described in Section~\ref{sec:targetdectionmici}. The measures were initialized randomly and then updated using the two different sampling techniques in optimization. Experiments were conducted in MATLAB using a desktop PC with Intel Xeon CPU 2.40 GHz processor and 16 GB RAM. The code was trivially parallelized and the running times are provided for relative comparisons only.

Table~\ref{table:meviruntime} lists the running times and the number of iterations until convergence for five runs of the two different optimization.  The algorithm was considered reaching convergence if the change of fitness is below $10^{-4}$. The ``ME'' in the table refers to the newly proposed sampling method that samples new measure element according to the counts of measure element used. The ``VI'' refers to the previously proposed method that samples new measure element by sorting the valid intervals of the measure elements \cite{du2016multiple}. It can be observed from Table~\ref{table:meviruntime} that the ``sampling according to measure element'' approach is in general faster in running time and converges in less iterations than the ``sort-by-valid-interval'' approach, mainly because the valid interval approach has to go through and evaluate the valid interval for all measure elements in each iteration, while the newly proposed ``measure element'' approach simply counts for the times a measure element was used in training once before the optimization starts. Besides, the fitness of the training data depends on the measure elements used, and the measure element that was most frequently used in training is updated most frequently in the ``measure element'' approach, which encourages the optimization to converge faster. The ROC curve performance are visibly very similar.

\begin{table}[h]
\centering
\caption{Running time (in seconds) and number of iterations until convergence for optimization schemes comparison. The standard deviation across five runs is noted in parentheses.}
\label{table:meviruntime}
\resizebox{0.49\textwidth}{!}{
\begin{tabular}{ |c|c|c|c|c|}
  \hline  
   & \multicolumn{2}{c|}{\textbf{ME}} & \multicolumn{2}{c|}{\textbf{VI}} \\ 
  \hline
  & NumIteration & Run Time &NumIteration & Run Time  \\
  \hline \hline
Run 1 & 213 & 79.7s & 3488 & 491.1s \\
\hline
Run 2 & 61 & 9.2s & 3536 & 490.4s \\
\hline
Run 3 & 256 & 37.0s & 2012 & 280.4s \\
\hline
Run 4 & 55 & 49.3s & 630 & 88.5s \\
\hline
Run 5 & 335 & 47.5s & 3231 & 448.0s \\
\hline
Summary &  & $\mathbf{44.5( 25.4)}$\textbf{s}  &  & $359.7(174.5)$s \\
\hline
\end{tabular}
}
\end{table}

\section{Conclusion}
\label{sec:conclusion}
This paper proposed  two novel Multiple Instance Choquet Integral models for classifier fusion and a novel Multiple Instance Choquet Integral Regression algorithm for remote sensing applications with uncertain and imprecise training labels. The newly proposed models successfully eliminated the variance parameter of previously proposed noisy-or fusion model, and extends for regression applications where the predictions are real-valued. In addition,  the newly proposed optimization scheme was able to significantly improve computation time. Experimental results show competitive performance of the proposed algorithms in target detection and crop yield prediction applications given real remote sensing data.

The current optimization scheme is based on evolutionary algorithm, which is effective but can be quite slow when the input image size increases. Alternative optimization schemes can be explored in future work, such as sampling measure elements more strategically using fitness values or develop a fast approximation approach using a message passing algorithm. The current regression model follows the standard ``primary instance'' assumption \cite{ray2001multiple}, which assumes that there is one primary instance in each bag that contributes to the regression label. However, this assumption may not necessarily be true in all data sets and it would be interesting to further relax the primary instance assumption. Currently, the normalized monotonic fuzzy measure is used in the proposed algorithms, yet it would be interesting to explore the binary fuzzy measure and/or bi-capacity Choquet integrals \cite{grabisch2005bi} as well as alternative integral approaches such as the Sugeno integral. Alternative features and remote sensing data sets as fusion sources can also be investigated.

% or


%\section*{Acknowledgment}


\ifCLASSOPTIONcaptionsoff
  \newpage
\fi



%\bibliographystyle{IEEEtran}

\bibliographystyle{IEEEtran} 
\bibliography{bibtex_entries}


\end{document}